%% file: 0-main.tex
\documentclass[10pt,twocolumn,letterpaper]{article}

\usepackage{cvpr}              
\makeatletter
\@namedef{ver@everyshi.sty}{}
\makeatother
\usepackage{tikz}

\usepackage{graphicx}
\usepackage{amsmath}
\usepackage{amssymb}
\usepackage{booktabs}
\usepackage{float}
\usepackage{blindtext}
\usepackage{multicol}
\usepackage{microtype}
\usepackage{wrapfig}
\usepackage{comment}
\usepackage{bm}
\usepackage{mathtools}
\usepackage[numbers]{natbib} 
\usepackage{setspace}
\usepackage{enumitem}
\usepackage{multirow}
\usepackage{makecell}
\usepackage{pifont}
\usepackage{caption}

\usepackage[ruled,norelsize]{algorithm2e}
\usepackage{algpseudocode}
\usepackage{algorithmicx}

\newtheorem{proof}{Proof}[section]

\usepackage{mdframed}
\definecolor{theoremcolor}{rgb}{0.94, 0.94, 0.94}
\definecolor{examplecolor}{rgb}{1, 1, 1.0}
\mdfsetup{
    backgroundcolor=theoremcolor,
    linewidth=0pt,
}
\usepackage{xfrac}
\usepackage{comment}
\newmdtheoremenv[linewidth=0pt,innerleftmargin=4pt,innerrightmargin=4pt]{defn}{Definition}
\newmdtheoremenv[linewidth=0pt,innerleftmargin=4pt,innerrightmargin=4pt]{prop}{Proposition}
\newmdtheoremenv[linewidth=0pt,innerleftmargin=4pt,innerrightmargin=4pt]{assump}{Assumption}
\newmdtheoremenv[linewidth=0pt,innerleftmargin=0pt,innerrightmargin=0pt,backgroundcolor=examplecolor]{example}{Example}
\newmdtheoremenv{corollary}{Corollary}
\newmdtheoremenv[linewidth=0pt,innerleftmargin=4pt,innerrightmargin=4pt]{theorem}{Theorem}
\newmdtheoremenv[linewidth=0pt,innerleftmargin=4pt,innerrightmargin=4pt]{lemma}{Lemma}
\DeclareMathAlphabet{\mathpzc}{OT1}{pzc}{m}{it}

\DeclareMathOperator*{\argmin}{arg\,min}

\newcommand{\abbr}{DDG}
\newcommand{\xt}{\mathbf{\tilde{x}}}
\newcommand{\x}{\mathbf{x}}
\newcommand{\DD}{\mathpzc{D}}
\newcommand{\PP}{\mathpzc{P}}
\newcommand{\MM}{\mathpzc{M}}

\usepackage{amssymb}
\usepackage{tikz}
\usepackage{colortbl}
\definecolor{Gray}{gray}{0.93}

\newcommand*{\circled}[1]{\lower.7ex\hbox{\tikz\draw (0pt, 0pt)%
    circle (.5em) node {\makebox[1em][c]{\small #1}};}}

\newlength\savewidth\newcommand\shline{\noalign{\global\savewidth\arrayrulewidth
  \global\arrayrulewidth 1pt}\hline\noalign{\global\arrayrulewidth\savewidth}}

\usepackage[pagebackref=true,breaklinks=true,colorlinks,bookmarks=false]{hyperref}
\hypersetup{linkcolor=[rgb]{0.7,0.1,0.1}}
\hypersetup{citecolor=[rgb]{0.4,0.15,0.95}}
\usepackage[capitalize]{cleveref}
\crefname{section}{Sec.}{Secs.}
\Crefname{section}{Section}{Sections}
\Crefname{table}{Table}{Tables}
\crefname{table}{Tab.}{Tabs.}
\newcommand\blfootnote[1]{%
  \begingroup
  \renewcommand\thefootnote{}\footnote{#1}%
  \addtocounter{footnote}{-1}%
  \endgroup
}

\def\cvprPaperID{} 
\def\confName{CVPR}
\def\confYear{2022}

\begin{document}

\title{\vspace{-4mm}Towards Principled Disentanglement for Domain Generalization}

\author{\fontsize{10pt}{\baselineskip}\selectfont Hanlin Zhang\textsuperscript{1,*}\ \ \ Yi-Fan Zhang\textsuperscript{2,*}\ \ \ Weiyang Liu\textsuperscript{3,4}\ \ \ Adrian Weller\textsuperscript{3,5}\ \ \ Bernhard Schölkopf\textsuperscript{4}\ \ \ Eric P. Xing\textsuperscript{1,6}\\[1.5mm]
\fontsize{10pt}{\baselineskip}\selectfont \textsuperscript{1}Carnegie Mellon University\ \ \ \  \textsuperscript{2}Chinese Academy of Science\ \ \ \  \textsuperscript{3}University of Cambridge\\
\fontsize{10pt}{\baselineskip}\selectfont \textsuperscript{4}Max Planck Institute for Intelligent Systems, Tübingen\ \ \ \ \textsuperscript{5}Alan Turing Institute\ \ \ \ \textsuperscript{6}MBZUAI\ \ \ \ \textsuperscript{*}Equal Contribution\\
}

\maketitle

\input{0-abstract}
\input{1-introduction}
\input{2-relatedwork} 
\input{4-methods}

\input{5-experiments}
\input{6-conclusions}
{\small
\bibliographystyle{ieee_fullname}
\bibliography{ref}
}

\newpage
\input{7-appendix}

\end{document}

%% file: 0-abstract.tex
\begin{abstract}
A fundamental challenge for machine learning models is generalizing to out-of-distribution (OOD) data, in part due to spurious correlations. To tackle this challenge, we first formalize the OOD generalization problem as constrained optimization, called \underline{\textbf{D}}isentanglement-constrained \underline{\textbf{D}}omain \underline{\textbf{G}}eneralization (DDG). We relax this non-trivial constrained optimization problem to a tractable form with finite-dimensional parameterization and empirical approximation. Then a theoretical analysis of the extent to which the above transformations deviates from the original problem is provided. Based on the transformation, we propose a primal-dual algorithm for joint representation disentanglement and domain generalization. In contrast to traditional approaches based on domain adversarial training and domain labels, DDG jointly learns semantic and variation encoders for disentanglement, enabling flexible manipulation and augmentation on training data. DDG aims to learn intrinsic representations of semantic concepts that are invariant to nuisance factors and generalizable across domains. Comprehensive experiments on popular benchmarks show that DDG can achieve competitive OOD performance and uncover interpretable salient structures within data.\blfootnote{Code repository: \url{https://github.com/hlzhang109/DDG}}
\end{abstract}

%% file: 1-introduction.tex
\section{Introduction}

Learning representations that can reflect intrinsic class semantics and also render strong invariance to cross-domain variation is of great significance to robustness and generalization in deep learning. Despite being empirically effective on many visual recognition benchmarks~\cite{russakovsky2015imagenet}, modern neural networks are still prone to learning shortcuts that stem from spurious correlations~\citep{Geirhos_2020}, resulting in poor out-of-distribution (OOD) generalization. To tackle this challenge, domain generalization (DG) has emerged as an increasingly important task where the goal is to learn invariant representations over source domains that are generalizable to distributions different from those seen during training \citep{MuaBalSch13}. 

\begin{figure}
  \centering  \includegraphics[width=0.46\textwidth]{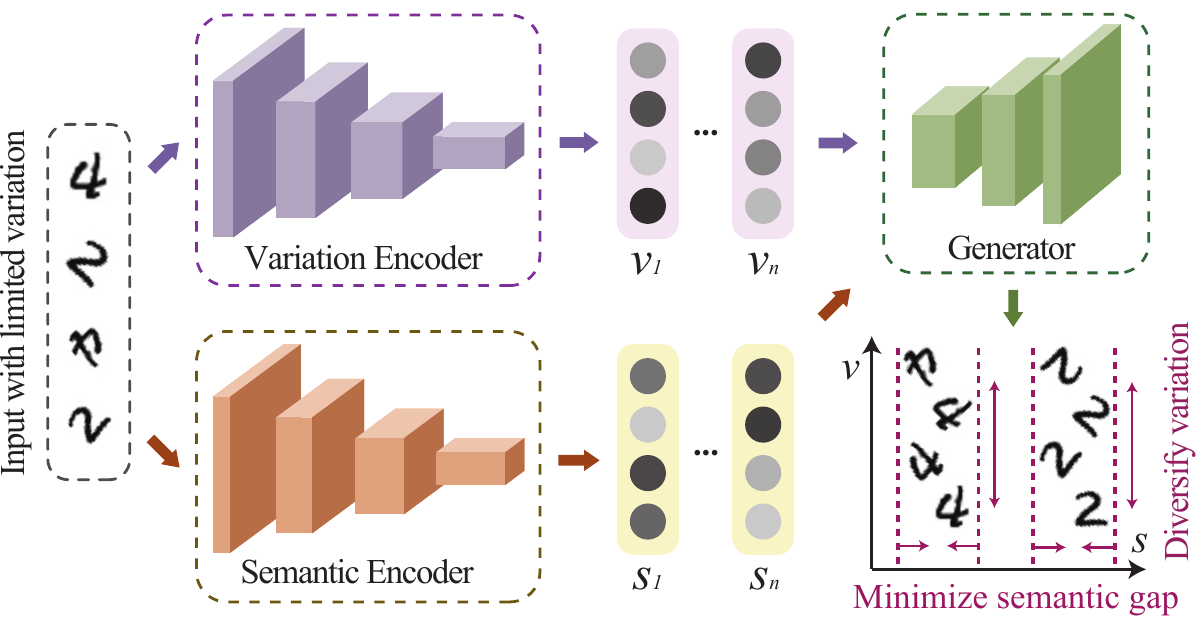}
  \vspace{-0.8mm}
  \caption{\footnotesize An illustration of \abbr~based on disentanglement of digit labels (semantics) and rotated angles (variation across domains). DDG seeks to minimize the semantic difference of the generated samples from the same class while diversifying the variation across source domains.}
  \vspace{-2.2mm}
  \label{fig:mnist_exp}
\end{figure}

In order to improve OOD generalization, efforts have been made from a diverse set of directions, such as domain adaptation \citep{2006Analysis, ben2010theory, ganin2016domain, zhao2018adversarial}, self-supervised learning \citep{carlucci2019domain, hendrycks2020pretrained}, causal inference \citep{RojSchTurPet18, SchJanPetSgoetal12,ZhaSchMuaWan13,peters2016causal}, invariant risk regularization \citep{arjovsky2019invariant, pmlr-v139-krueger21a,lu2021nonlinear}, angular alignment regularization \citep{chen2020angular,liu2017deep,liu2021orthogonal}, distributionally robust optimization \citep{ben2009robust} and data augmentation \citep{bai2020decaug, zhang2019unseen, volpi2018generalizing}. However, having no access to target domain data poses great challenges. Two main lines of research seek to address this. First, training domain labels are assumed to be available in \citep{ganin2016domain, li2018deep, robey2021model, arjovsky2019invariant, sagawa2020distributionally} such that the divergence to different domains can be minimized. However, these domain labels are often impractical or prohibitively expensive to obtain \citep{hashimoto2018fairness}. Moreover, it is non-trivial to minimize domain divergence with domain adversarial training which is notoriously hard to converge \citep{roth2017stabilizing}. The second line of works tries to model the cross-domain distribution shifts and capture the semantic invariance \citep{rosenfeld2020risks, kamath2021does, wang2021toward}. However, it has been found in \citep{gulrajani2021in} that this goal can be very difficult to achieve. What makes the problem even more challenging is the inconsistency of the evaluation protocol. Surprisingly, \citep{gulrajani2021in} shows that even the standard empirical risk minimization could outperform many recently proposed models under certain conditions. Motivated by these challenges, we aim to disentangle variations and semantics in a principled way, and then verify the effectiveness of our method under a consistent evaluation protocol~\cite{gulrajani2021in}.

A key desideratum for DG is to ensure the invariance of learned representations to all possible inter-class variations. Therefore, our intuition is to first diversify the inter-class variation by modeling potential seen or unseen variations, and then minimize the discrepancy of the inter-class variation on a representation space where the target is to predict semantic labels. To this end, we first formalize distribution shifts and invariance based on disentanglement. 
Concretely, we formulate the disentanglement between class semantics and both intra- and inter-domain variations as constraints to the DG problem. Then we propose a novel framework called \underline{\textbf{D}}isentanglement-constrained \underline{\textbf{D}}omain \underline{\textbf{G}}eneralization (DDG). An illustration of DDG is given in Fig.~\ref{fig:mnist_exp}. When the semantic (i.e. the labels of digits) and variation factors (i.e. rotating angles $-60^{\circ}, 0^{\circ}, 60^{\circ}, 120^{\circ}$) are well disentangled, we expect that learned representations can be effectively constrained to be invariant to inter-class variation. In order to achieve such a non-trivial goal, we first derive a constrained optimization problem and then propose a principled algorithm based on primal-dual iterations to solve it. To understand how well the transformed solution approximates the solution to the original problem, we provide comprehensive theoretical guarantees for the parameterization gap and empirical gap. 
We also verify the empirical effectiveness of DDG by showing that it can consistently outperform current popular DG methods by a considerable margin. 

As a useful side product, DDG simultaneously obtains an automated, domain-agnostic data augmentation network based on learned disentangled representations. This requires no usage of domain-specific knowledge or gradient estimation~\citep{bai2020decaug, volpi2018generalizing}. The intuition why such a data augmentation network is useful comes from the fact that the learned variation encoder can well approximate some of the intrinsic intra- and inter-domain variations. It also serves as \textit{feature removal} since more training examples augmented by specific variation factors lead to more invariant representations for those variations. Moreover, the increased diversity of source domain data improves the likelihood that an unseen distribution lies within the convex hull of source domains \citep{albuquerque2019generalizing}. For example, in Fig.~\ref{fig:mnist_exp}, the original dataset can be augmented via a learned manipulator by composing a diverse combination of semantic and variation factors. Such a disentanglement can be a good predictor for OOD generalization according to \citep{dittadi2021on}. We highlight the following advantages of DDG:

\vspace{0.5mm}
\begin{itemize}[leftmargin=*,nosep]
\setlength\itemsep{0.3em}
\item DDG adopts a principled constrained learning formulation based on disentanglement, yielding rigorous theoretical guarantees on the empirical duality gap. 
\item Our algorithm is conceptually simple yet effective. DDG promotes semantic invariance via a constrained optimization setup. This is done without the usage of adversarial training and domain labels. Moreover, there is no additional computational overhead for modeling variations. 
\item Our framework can be viewed as a controllable and interpretable data generation paradigm for DG. Data manipulation under domain transformation can be challenging in settings where domain-specific signals like image styles vary greatly across domains, constituting more complicated superficial variation factors. Yet \abbr~can uncover salient structure within data by imposing constraints on the semantic and variation factors. 
\item Comprehensive experiments are conducted under a consistent evaluation protocol to verify the effectiveness of \abbr. We show that \abbr~is able to produce interpretable qualitative results and achieve competitive performance on a number of challenging DG benchmarks including RotatedMNIST, VLCS, PACS and WILDS.
\end{itemize}

\label{sec:intro}

%% file: 2-relatedwork.tex
\vspace{-0.5mm}
\section{Related Work}
\vspace{-0.5mm}

\textbf{Domain Generalization.} Domain/Out-of-distribution generalization \cite{MuaBalSch13} aims to learn representations that are invariant across domains so that the model can extrapolate well in unseen domains. 
Invariant Risk Minimization (IRM) \cite{arjovsky2019invariant}, which extends \citep{peters2016causal}, and its variants \cite{ahuja2020invariant, pmlr-v139-krueger21a, lu2021nonlinear} are proposed to tackle this challenge. However, IRM entails challenging bi-level optimization and can fail catastrophically unless the test data are sufficiently similar to the training distribution \citep{rosenfeld2020risks}. 
DG via domain alignment \cite{albuquerque2019generalizing,MuaBalSch13,chen2020angular} aims to minimize the difference between source domains for learning domain-invariant representations. The motivation is straightforward: features that are invariant to the source domain shift should also be robust to any unseen target domain shift. The main difference is that we propose to learn invariant representations by reconstructing images from various domains and class semantics to simulate variations and minimize domain divergence. PAC constrained learning~\cite{chamon2020probably, chamon2020empirical} is adopted for modeling cross-domain variations under domain transformation in MBDG \citep{robey2021model}. We highlight several major differences between our approach and MBDG below: (1) \abbr~imposes weaker assumptions;
(2) MBDG consumes additional domain labels, which are often hard to obtain in many applications, while \abbr~does not;
(3) \abbr~enforces invariance constraints via parameterizing semantic and variation encoders, which does not belong to a model-based approach. In contrast, MBDG requires a pre-trained domain transformation model (\eg, CycleGAN) during training. Appendix~\ref{app:exp_results} provides the detailed comparison to MBDG.

\textbf{Disentangled Representation Learning.} The goal of disentangled representation learning is to model distinct and explanatory factors of variation in the data \citep{bengio2013representation,scholkopf2021toward}. \cite{dittadi2021on} shows that disentanglement is a good predictor for out-of-distribution (OOD) tasks. \cite{Ruichu2019Learning} proposes to disentangle the semantic latent variables and the domain latent variables for stronger generalization performance in domain adaptation. \cite{montero2021the} shows that existing disentangled learning models are not sufficient to support compositional generalization and extrapolation while hypothesizing that the richness of the training domain matters more. However, previous works \citep{higgins2016beta, kim2018disentangling, chen2018isolating} are limited to single-dimensional latent codes and developed with different purposes like generation and interpretability. Thus they are hard to scale well beyond toy datasets and adapt to complicated DG tasks \citep{montero2021the}. In contrast, we harness the disentangled effects to learn invariant representations for realistic OOD generalization tasks. 

\textbf{Data Augmentation.} The diversity of the training distribution is of great importance in improving DG performance~\citep{albuquerque2019generalizing, gulrajani2021in, hendrycks2020pretrained, tu2020empirical, zhang2019unseen}. Data augmentation is an effective way to increase data diversity \citep{xie2020unsupervised} and it can therefore improve OOD generalization as well as robustness to spurious correlations~\citep{kaushik2019learning, bai2020decaug}. In particular, \cite{chalupka2015visual} devises an active learning scheme to learn causal manipulations on images, which enriches the dataset from observational data and improves generalization on both causal and predictive learning tasks. In contrast, \abbr~seeks to learn underlying causal features by approximating the data manipulation function. This is done without a task-specific metric to differentiate the augmented data and the oracle. Our work introduces a simple yet effective approach for augmenting training data, which reinforces the importance of data diversity in DG.

\textbf{Fairness.} Fairness research~\citep{hashimoto2018fairness, dwork2012fairness, madras2018learning} aims to develop a model that performs well under group assignments according to some fairness criteria for addressing the underperformance in minority subgroups. Learning fair representations can be naturally translated to a constrained optimization problem~\citep{chamon2020probably, chamon2020empirical, kusner2017counterfactual}. There are also exchanging lessons between algorithmic fairness and domain generalization~\citep{creager2021environment}, showing that both fields are optimizing similar statistics for common goals. \abbr~well aligns with the formulation and goals of \textit{fairness without demographics} \citep{hashimoto2018fairness} and has the potential to improve context-specific fairness without prior knowledge about domains or demographics.

%% file: 4-methods.tex
\section{Disentanglement-constrained Optimization for Domain Generalization}

\textbf{Notations.} We consider a classification problem from feature space $\mathcal{X} \in \mathbb{R}^d$ to label space $\mathcal{Y} \in \{0,1\}$ where $(X, Y) \sim \mathbb{P}(X,Y)$. The infinite-dimensional functional space and finite-dimensional hypothesis space are denoted as $\mathcal{F}$ and $\mathcal{H} \subseteq \mathbb{R}^{p}$, respectively. The parameterized latent spaces for semantic and variation factors are denoted as $\mathcal{S}$ and $\mathcal{V}$, respectively. $\xt$ denotes a different sample with $\x$ from the training distribution $\mathbb{P}$. $d(\cdot,\cdot)$ denotes a distance metric over $\mathcal{X}\times\mathcal{X}$. 

\textbf{Problem setting.} Suppose we observe a dataset denoted as $\left\{\left(\boldsymbol{x}_{i}, y_{i} \right)\right\}_{i=1}^{n} \subset \mathcal{D}_{data}$ where $(\boldsymbol{x}_{i}, y_{i})$ is a realization of random vector $(X, Y)$ with support $(\mathcal{X} \times \mathcal{Y})$. We consider a set of domains $\left\{d_i\right\}_{i=1}^{n_d} \subset \mathcal{D}$ of size $n_d$, where each domain corresponds to a distinct data distribution $D^{d_i}$ over some input and label space. The set of domains $\mathcal{D}$ is partitioned into multiple training domains $\mathcal{D}_{S} \subset \mathcal{D}$ and a test domain $\mathcal{D}_{U} \in \mathcal{D}$ which is inaccessible during training.

Our formulation generally follows prior works of PAC constrained learning~\citep{chamon2020probably, chamon2020empirical, robey2021model, robey2021adversarial}, but we use a more flexible parameterization and derive a new algorithm to solve the resulting constrained optimization problem. More specifically, we emphasize that \abbr, motivated by an analysis of the multi-source domain adaptation upper bound (Appendix \ref{app:da_bound}), requires no domain labels and pre-trained domain transformation models during training. \abbr~can also be trained in an end-to-end manner, yielding a more flexible and potentially better solution.

\vspace{0.55mm}
\subsection{Formulation}

The basic idea of \abbr~is to learn disentangled representations by imposing invariant constraints in the semantic space $\mathcal{S}$ and variation space $\mathcal{V}$. Such a disentanglement can also be applicable for augmenting the training data so that the learned representations can be more invariant to both inter- and intra-domain variations. To formalize this, we begin by introducing some necessary definitions and assumptions.

\vspace{0.2mm}
\begin{defn}{(Invariance based on disentanglement)}. Given a decoder $D: \thickmuskip=2mu \medmuskip=2mu \mathcal{S} \times \mathcal{V} \rightarrow \mathcal{X}$, a semantic featurizer $f_s$ is invariant if for all domains $\thickmuskip=2mu \medmuskip=2mu d_i \in \mathcal{D}$ and a variation featurizer $f_v$, $\thickmuskip=2mu \medmuskip=2mu\x = D(f_s(\x;\theta), f_v(\xt;\phi))$ holds almost surely when $\thickmuskip=2mu \medmuskip=2mu\x, \xt \sim P(X)$.
\end{defn}
\vspace{0.2mm}

The property enforces the invariance of the original input $\x$ and the one $\thickmuskip=2mu \medmuskip=2mu D(f_s(\x;\theta), f_v(\xt;\phi))$ that reconstructs jointly from semantic and variation latent spaces when semantic factors remain constant while variation factors vary.

\vspace{0.2mm}
\begin{assump}{(Domain shift based on disentanglement)} Denote $f_s(\x;\theta)$ as the semantic factor of input $\x$ and $f_v(\xt;\phi)$ as the variation factor of any other one $\xt$. Similar to the covariate shift assumption \citep{shimodaira2000improving}, we assume the domain/distribution shift stems from the variation of the marginal distribution $P(X)$ and the following invariance condition holds with the proposed $f_s, f_v$ and $D$

\vspace{-4.5mm}
\begin{equation}
\small
    \!\!\!\!P(Y\!=y|X\!=
    \x)= P(Y\! = y| X\! =\! D(f_s(\x;\theta), f_v(\xt;\phi)). 
\end{equation}
\vspace{-4.5mm}
\label{assump}
\end{assump}
\vspace{0.2mm}

This assumption shows that the prediction depends only on the semantic factor $f_s(\x;\theta)$ regardless of the variation one $f_v(\xt;\phi)$. It also subsumes as a special case the domain shifts based on domain labels, \ie, $\thickmuskip=2mu \medmuskip=2mu P(Y=y|X=\x) = P(Y^d = y| X^d = G(X=\x,d))$ given a domain transformation model \citep{robey2021model, robey2020modelbased}, since our variation factors includes both inter- and intra-domain variations.

Note that the above assumptions follow the formulation in \citep{robey2021model}. To elaborate this, we introduce the notion of invariance based on a decoder $D$ taking as input the disentanglement results $f_s(\x;\theta)$ and $f_v(\xt;\phi)$. In practice, $D$ can be parameterized as a pre-trained model or a trainable component $D_\psi$ updated in the primal step as in our implementation.
\newpage

\begin{assump}{(Regularity conditions)}
The loss function $\ell$ and distance metric $d$ are convex, non-negative, B-bounded. $\ell$ is a $L_\ell$-Lipschitz function, the distance metric is also a $L_d$-Lipschitz function. 
\label{assump:regularity}
\end{assump}
\vspace{-2.1mm}

\begin{assump}{(Feasibility)}
There exist semantic and variation featurizers $f_{s}, f_v \in \mathcal{F}$ such that $\mathcal{L}_{con}(f_s, f_v) < \gamma - m = \max \{L_\ell\epsilon_s, L_d\epsilon_g\} = \gamma - m$ with $\epsilon$-parameterization. 
\label{assump:feasibility}
\end{assump}
\vspace{-2.1mm}

\begin{defn}{(Domain generalization problem).} Similar to prior works \citep{wang2021toward, sinha2018certifying, sagawa2020distributionally, robey2021model}, we formulate domain generalization as a minimax optimization problem, optimizing the worst-domain risk over the entire family of domains $\mathcal{D}$
\vspace{-1mm}
\begin{equation}
\small
    \underset{f_s \in \mathcal{F}}{\min} \max_{d \in \mathcal{D}} \mathbb{E}_{\mathbb{P}(X, Y)} \ell(f_s(D(X, d)), Y).
\label{eq:ori_dg}
\end{equation}
\vspace{-4.4mm}
\end{defn}

The above formulation (\ref{eq:ori_dg}) requires the availability of domain labels and is hard to optimize. However, the domain labels are expensive or even impossible to obtain in part due to privacy and fairness issues \citep{hashimoto2018fairness}. Therefore, under the disentanglement-based invariance and domain shift assumptions, we constrain the model to be invariant with respect to variation factors, then the problem is converted to an inequality-constrained optimization problem:

\begin{defn}(Constrained domain generalization problem)
Given a fixed margin $\gamma > 0$, with Assumption \ref{assump:feasibility} and enforcing the invariance on the semantic featurizer $f_s$, we transform the vanilla formulation Eq.~(\ref{eq:ori_dg}) to the following inequality-constrained optimization
\vspace{-0.85mm}
\begin{equation}
\small
\begin{aligned}
& \PP^{\star} \triangleq \underset{f_s\in\mathcal{F}}{\operatorname{min}} \ \mathcal{L}(f_s) \triangleq \mathbb{E}_{\mathbb{P}(X, Y)} \ell(f_s(X), Y), \\ 
&\textnormal{s.t.}~
 d(\x, D(f_s(\x;\theta),f_v(\xt;\phi))) \le \gamma, \  \textnormal{a.e.}~\x, \xt \sim \mathbb{P}(X). 
\end{aligned}
\label{eq:relax_dg}
\end{equation}
\end{defn}
\vspace{0.2mm}

One intriguing property of Eq. (\ref{eq:relax_dg}) is that learning with inequality constraints does not produce additional sample complexity overhead under some regularity conditions on the loss function $\ell$ \cite{chamon2020probably}. However, it is difficult to satisfy the strictness and provide theoretical guarantees for learning in practical cases. In the following section, with the parameterization and saddle-point condition, we can relax the invariant constraint and obtain a version that is amenable to a provable PAC learning framework. 

\subsection{Parameterization}
We first discuss how to parameterize the learnable components in \abbr. The DG problem (Eq.~\eqref{eq:relax_dg}) yields an infinite-dimensional optimization. A \emph{de facto} way to enable tractable optimization is using finite-dimensional parameterization of $\mathcal{F}$ like neural networks \citep{hornik1989multilayer} or reproducing kernel Hilbert spaces (RKHS) \citep{berlinet2011reproducing}. To further discuss the parameterization gap, we formalize the approximation power of such parameterization by the following definition of $\epsilon$-parameterization.

\begin{defn}{($\epsilon$-parameterization)} Let $\mathcal{H} \subseteq \mathbb{R}^{p}$ be a finite-dimensional parameter space. For $\epsilon>0$, a function $h: \mathcal{H} \times \mathcal{X} \rightarrow \mathcal{Y}$ is an $\epsilon$-parameterization of $\mathcal{F}$ if for each $f_s, f_v \in \mathcal{F}$, there exist parameters $\theta, \phi \in \mathcal{H}$ such that
\vspace{-1mm}
\begin{equation*}
\small
\begin{aligned}
& \mathbb{E}_{\x\sim P(X)}\|h_s(\x;\theta)-f_s(\x)\|_{\infty} \leq \epsilon_s, \\
& \mathbb{E}_{\x\sim P(X)}\|h_v(\x;\phi)-f_v(\x)\|_{\infty} \leq \epsilon_v, \\
& \mathbb{E}_{\x,\xt \sim P(X)}\|D(h_s(\x;\theta), h_v(\xt;\phi))-D(f_s(\x), f_v(\xt))\|_{2} \leq \epsilon_g. \\
\end{aligned}
\end{equation*}
\end{defn}

With the help of $\epsilon$-parameterization, tractable optimization can be performed over finite-dimensional parameterized space. Note that a regularity condition about $D$ is introduced to allow \abbr~to faithfully reconstruct inputs under finite-dimensional parameterization. With the above formulation and to provide guarantees for the DG problem, we consider a corresponding saddle-point problem as follows:
\begin{equation}
\small
    \DD_\epsilon^*(\gamma) \triangleq
    \max\limits_{\lambda} \min\limits_{\theta, \phi\in\mathcal{H}}
    \mathcal{L}(\theta) + \lambda \mathcal{L}_{con}(\theta, \phi), 
    \label{eq:param}
\end{equation}
where the constraint-related risk is defined as
\begin{equation*}
\small
\begin{aligned}
    \mathcal{L}_{con}(\theta, \phi) = \mathbb{E}_{\x, \xt \sim P(X)}[d(\x, D(h_s(\x;\theta),h_v(\xt;\phi))) - \gamma]. \\
\end{aligned}
\end{equation*}
The challenge for the parameterized problem (\ref{eq:param}) is the inaccessibility of the ground truth data distribution $\mathbb{P}(X,Y)$. To address it, we resort to a corresponding empirical dual problem using finite $n$ empirical training samples:
\begin{equation}
\small
\begin{aligned}
    \DD_{\epsilon, n}^{*}(\gamma) & \triangleq
    \max\limits_{\lambda} \min\limits_{\theta, \phi\in\mathcal{H}} L(\theta, \phi, \gamma) \triangleq
    \hat{\mathcal{L}}(\theta) + \lambda \hat{\mathcal{L}}_{con}(\theta, \phi) \\
    &= \max\limits_{\lambda} \min\limits_{\theta, \phi\in\mathcal{H}} \sum\limits_{i=1}^{n} \ell(f_s(\mathbf{x_i}), y_i) + \\ & \lambda \sum\limits_{i=1}^{n}\sum\limits_{j\neq i}^{n} \left[ d\left(\mathbf{x_i}, D\left(h_{s}(\mathbf{x_i} ; \theta), h_{v}(\mathbf{x_j}; \phi)\right)\right)-\gamma \right],
\label{eq:emp_dual_gap}
\end{aligned}
\end{equation}
which gives us the final optimization objective for DDG. Compared to the previous optimization problems, this is much easier and more tractable to solve.

\subsection{Algorithm}
Motivated by the above analysis, we use a primal-dual algorithm for efficient optimization \citep{chamon2020probably, chamon2021constrained, robey2021model, robey2021adversarial}. 
The algorithm alternates between optimizing $\theta$ (and/or $\phi$) via minimizing the empirical Lagrangian with fixed dual variable $\lambda$ and updating the dual variable according to the minimizer:
\begin{equation}
\small
\begin{aligned}
& \theta^{(t+1)} \leftarrow \argmin_{\theta} L(\theta^{(t)}, \phi^{(t)}, \gamma) + \rho, \\
& \phi^{(t+1)} \leftarrow \argmin_{\phi} L(\theta^{(t)}, \phi^{(t)}, \gamma) + \rho, \\
& \lambda^{(t+1)} \leftarrow \max \left\{\left[\lambda^{(t)}+\eta_{2}\hat{\mathcal{L}}_{\text {con}}\right], 0\right\},
\end{aligned}
\end{equation}
where the $\eta_2$ denotes the learning rate of the dual step.

The primal-dual iteration has clear advantages over stochastic gradient descent in solving constrained optimization problems. Specifically, it avoids introducing extra balancing hyperparameters. Moreover, it provides convergence guarantees once we have sufficient iterations and a sufficiently small step size. We refer readers to \citep{chamon2021constrained, robey2021model} for more in-depth and complete discussions of related conditions and convergence bounds. 

One intriguing property of disentanglement is that it can be applicable for augmenting the training data. Based on this, \abbr~approximates a manipulator function by learning ``hard'' data points from fictitious target distributions for promoting invariance and improving generalization.

We give the detailed procedure of our \abbr~learning algorithm in Algorithm~\ref{algo:main}, where $\ell$ denotes the cross entropy loss and $\oplus$ denotes the concatenation in the implementation. We also use $l_1$ norm as the distance metric $d$ in the experiments.

\begin{algorithm}[t]
\caption{DDG: Disentanglement-constrained Optimization for Domain Generalization}
\SetAlgoLined
\KwIn{$\mathcal{D}_S = \{ \left(\mathbf{x_1}, y_1 \right), ... , \left(\mathbf{x_n}, y_n \right) \}$, batch size $B$, primal and dual learning rate $\eta_1,\eta_2$, Adam hyperparameters $\beta_1$, $\beta_2$, initial coefficients $\lambda$, margin $\gamma$}
\textbf{Initial}: Parameters of DDG (i.e. parameters $\theta$, $\phi$ and $\psi$ for semantic encoder $h_s$, variation encoder $h_v$ and decoder $D$.)

\Repeat{$\theta$ is converged or $\mathcal{D}_S = \emptyset$}{
    \For{$i,j = 1, \dots, B$, $i \neq j$}
    {$\mathcal{L}^i_{\text {con }}=\max\left\{\left\|\mathbf{x_i}-D\left( h_s(\mathbf{x_i}) \oplus  h_v(\mathbf{x_j})\right)\right\|_{l_{1}}-\gamma, 0\right\}$ 
    \begin{flushleft}
     $\mathcal{L}^i_{\text{ERM}} = \ell(h_s(\mathbf{x_i}), y_{i})$\\
     $\mathcal{L}_{i} = \mathcal{L}_{\text{ERM}}^{i} + \lambda \mathcal{L}_{\text {con}}^{i}$
    \end{flushleft}
     \If{Data Augmentation}{
    
    $\mathbf{x}^* = D\left( h_s(\mathbf{x_i}) \oplus  h_v(\mathbf{x_j})\right)$ \\
    $\mathcal{L}^*_{\text{ERM}} = \ell(h_s(\mathbf{x}^*), y_{i})$\\
    $\mathcal{L}_{i}=\mathcal{L}_{i} + \mathcal{L}^*_{\text{ERM}}$
}
    }
    \textbf{Primal step}
    
    $\theta \leftarrow \operatorname{Adam}\left(\frac{1}{B} \sum_{i=1}^{B} \mathcal{L}_{i}, \theta, \eta_1, \beta_{1}, \beta_{2}\right)$ \\
    $\phi \leftarrow \operatorname{Adam}\left(\frac{1}{B} \sum_{i=1}^{B} \mathcal{L}_{con}^{i}, \phi, \eta_1, \beta_{1}, \beta_{2}\right)$ \\  
    \If{Training $D$}{$\psi \leftarrow \operatorname{Adam}\left(\frac{1}{B} \sum_{i=1}^{B} \mathcal{L}_{con}^{i}, \psi, \eta_1, \beta_{1}, \beta_{2}\right)$}  
    \textbf{Dual step} 
    
    $\lambda\leftarrow\max \left\{ \left[\lambda+\eta_2\frac{1}{B}\sum_{i=1}^B\mathcal{L}^i_{\text{con}}\right], 0\right\}\;$
    } 
\label{algo:main}
\end{algorithm}
\subsection{Theoretical Insights and Guarantees}
In this subsection, we provide a comprehensive analysis of the statistical guarantees of our solution (Eq.~(\ref{eq:emp_dual_gap})). In order to derive the bound on the final empirical duality gap, we start by proving two lemmas on the parameterization gap and empirical gap. Specifically, they elaborate the corresponding approximation gaps of two transformations (i.e. Eq.~(\ref{eq:param}) and (\ref{eq:emp_dual_gap})) in above sections.

We first discuss the gap between the finite-dimensional model parameterization (e.g. neural networks) and the model over infinite functional space $\mathcal{F}$.

\begin{lemma}{(Parameterization gap)}\label{bound:param_gap} 
With Assumption \ref{assump:regularity} about $\ell$ and $d$, the gap between optimum of a statistical problem and its finite dimensional, deterministic version $\DD_{\varepsilon}^{\star}(\gamma)-\PP^{\star}$ can be bounded as
\begin{equation}\label{main_para_gap}
\small
0 \le \DD_{\varepsilon}^{\star}(\gamma) - \PP^{\star} \le \left(1+|\lambda_{p}^{\star}|\right)  \max \left\{L_{\ell}\epsilon_s, L_{d}\epsilon_g\right\},
\end{equation}
where $\lambda_{p}^{\star}$ is the dual variable with a tighter constraint $\gamma - \max \left\{L_{\ell}\epsilon_s, L_{d}\epsilon_g\right\}$ in Eq. (\ref{eq:relax_dg}).
\end{lemma}

The upper bound indicates that the parameterization gap is dominated by both the semantic function parameterization and the reconstruction-based transformation on perturbed inputs, which makes intuitive senses and also emphasizes the important role of disentanglement.

Then we compare Eq.~\eqref{main_para_gap} to the parameterization gap of MBDG which is shown as follows:
\begin{equation}
\small
    0 \leq \DD_{\epsilon}^{\star}(\gamma) - \PP^{\star} \leq (1+|\lambda_{p}^{\star}|) \max \{L_{\ell}, L_{d}\} \epsilon_s,
    \label{bound:model-dg}
\end{equation}
and the parameterization gap in \citep{chamon2020empirical}:
\begin{equation}
\small
    0 \leq \DD_{\epsilon}^{\star}(\gamma) - \PP^{\star} \leq (1+|\lambda_{p}^{\star}|) L_{\ell} \epsilon_s.
\end{equation}
From the comparison, we notice that our formulation and analysis are closely connected to Proposition~1 (with $m=1$) in \citep{chamon2020empirical}. We shall also see that in a perfect case where representations are well disentangled, \ie, $\epsilon_g \rightarrow 0$, our bound will become $\left(1+|\lambda_{p}^{\star}|\right) L_{\ell} \epsilon_{s}$. We note that this bound is strictly tighter than that in Eq. (\ref{bound:model-dg}).

In practice, we approximate the expectation by its empirical average. By the classical VC-dimension bound, the following bound on the empirical gap holds:

\begin{lemma}{(Empirical gap)} Denote $d_{VC}$ as the VC-dimension of the hypothesis class $\mathcal{H}_\theta$. Assume that $\ell$ and $d$ obey the regularity condition in Assumption \ref{assump:regularity}. Then given $n$ samples, with probability $1-\delta$, we can upper bound the deviation $|\DD^{\star}_{\epsilon}(\gamma)-\DD_{\varepsilon, n}^{\star}(\gamma)|$ with
\begin{equation}
\small
\!\!\!\!\left|\DD_{\epsilon}^{\star}(\gamma)-\DD_{\varepsilon, n}^{\star}(\gamma)\right| \le 2B \sqrt{\frac{1}{n}\left[1+\log \left(\frac{4(2 n)^{d_{v c}}}{\delta}\right)\right]}.
\end{equation}
\label{bound:emp_gap}
\vspace{-3mm}
\end{lemma}

\begin{figure*}[t!]
    \centering
    \includegraphics[width=0.99\textwidth]{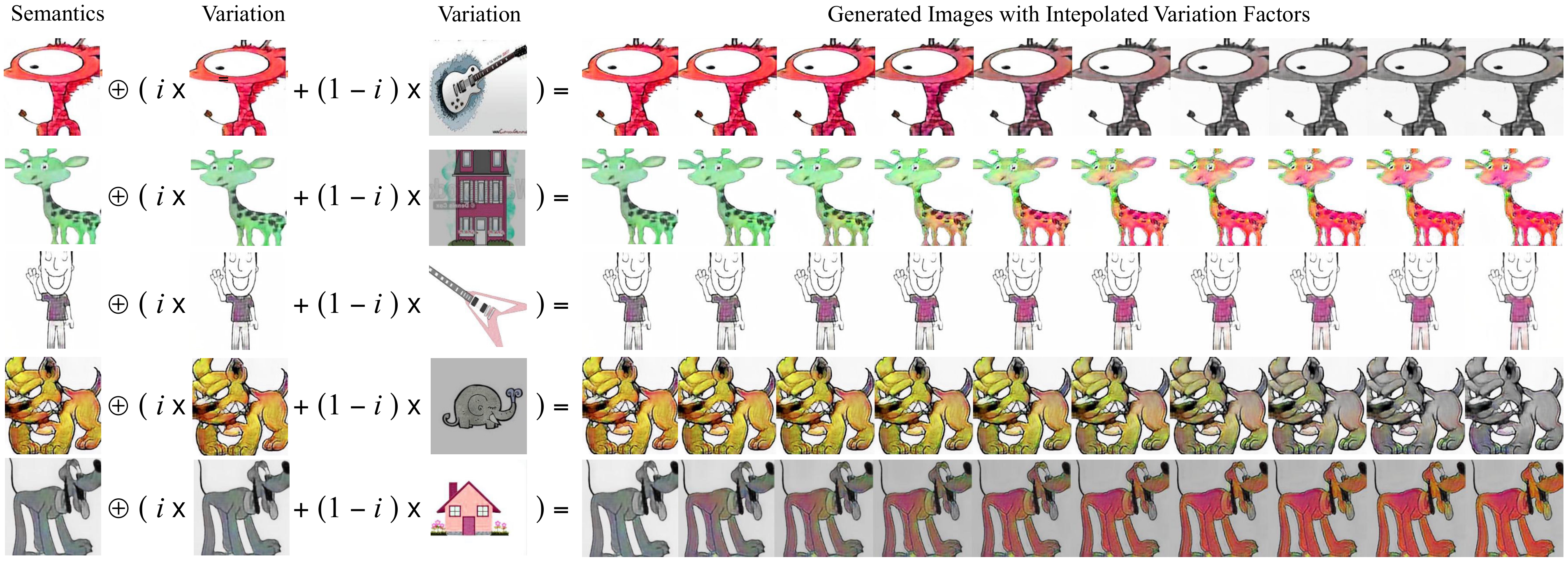}
    \caption{\footnotesize Interpolation disentanglement results. Different proportions of variation factors are mixed to generate the image by varying $i \in \{1.0, 0.9, \dots, 0.1\}$.} 
    \label{fig:interpolation}
\end{figure*}

With the above heavy lifting, we start deriving the empirical duality gap, which is our ultimate goal of the theoretical analysis. The empirical duality gap includes the above two components. Combining the above bounds on two gaps, we can bound the deviation between $\PP^{\star}$ and $\DD_{\epsilon, n}^*(\gamma)$ (\ie, $|\PP^{\star}-\DD_{\epsilon, n}^*(\gamma)|$) under some mild conditions.

\begin{theorem}{(Empirical duality gap)} When Assumption \ref{assump:regularity} holds, by denoting $\max \left\{L_{\ell} \epsilon_{s}, L_{d} \epsilon_{g}\right\}$ as $m$, we have

\vspace{-4mm}
\begin{equation}
\small
\left|\PP^{\star}-\DD_{\varepsilon, n}^{\star}(\gamma)\right| \leq (1+|\lambda|) m + \mathcal{O}(\sqrt{\frac{\log(n)}{n}}).
\end{equation}
\label{bound:emp_dual_gap}
\vspace{-4mm}
\end{theorem}

The final bound tells us that the quality of the empirical, dual approximation of the primal problem is determined by the sample size, the hardness of the learning problem, and the richness of parameterization. The proof can be easily shown using triangle inequality as in Appendix \ref{app:emp_duality_proof}. As suggested by Theorem~\ref{bound:emp_dual_gap}, we can improve the performance of our algorithm by using neural networks with larger capacity or training our model with more data.

%% file: 5-experiments.tex
\section{Experiments}
\label{sec:exp}
\textbf{Datasets.} We consider the following four datasets: Rotated MNIST \citep{ghifary2015domain}, PACS \citep{li2017deeper}, VLCS \citep{torralba2011unbiased} and WILDS \citep{koh2021wilds} to evaluate \abbr~against previous methods. We include the visualization of datasets in Appendix \ref{app:dataset}. 

\textbf{Rotated MNIST} \citep{ghifary2015domain} consists of 10,000 digits in MNIST with different rotated angles $d$ such that each domain is determined by the degree $d \in \{ 0, 15, 30, 45,
60, 75 \}$.

\textbf{PACS} \citep{li2017deeper} includes 9, 991 images with 7 classes $y \in  \{$ dog, elephant, giraffe, guitar, horse, house, person $\}$ from 4 domains $d \in$ $\{$ art, cartoons, photos, sketches $\}$. 

\textbf{VLCS} \citep{torralba2011unbiased} is composed of 10,729 images, 5 classes $y \in \{$bird, car, chair, dog, person$\}$ from domains $d \in \{$Caltech101, LabelMe, SUN09, VOC2007$\}$. 

\textbf{Camelyon17-WILDS} \citep{koh2021wilds, 8447230} is about tumor detection in tissues. This dataset is composed of 455,954 images from 5 different medical centers/domains in total, which defines a binary classification problem about whether the patch image contains a tumor or not.

\textbf{Baselines.} We compare our model with ERM \citep{vapnik1998statistical}, IRM \citep{arjovsky2019invariant}, GDRO \citep{sagawa2020distributionally}, Mixup \citep{yan2020improve}, MLDG \citep{li2018learning}, CORAL \citep{sun2016deep}, MMD \citep{li2018domain}, DANN \citep{ganin2016domain}, CDANN \citep{li2018deep}, AugMix \citep{hendrycks*2020augmix}.

All the baselines in DG tasks are implemented using the codebase of Domainbed \citep{gulrajani2021in}. We adapt AugMix for ablation studies using the official implementations as indicated in \citep{hendrycks*2020augmix}. The two-sample classifier, implemented as a RBF kernel SVM using Scikit-learn, is used for calculating the generalization error for $\mathcal{A}$-distance.

\textbf{Hyperparameter search.} Following the experimental settings in \citep{gulrajani2021in}, we conduct a random search of 20 trials over the hyperparameter distribution for each algorithm and test domain. Specifically, we split the data from each domain into $80\%$ and $20\%$ proportions, where the larger split is used for training and evaluation, and the smaller one is for selecting hyperparameters. We repeat the entire experiment twice using different seeds to reduce the randomness.
Finally, we report the mean over these repetitions as well as their estimated standard error.

\textbf{Model selection.} The model selection in domain generalization is intrinsically a learning problem, and we use test-domain validation, one of the three selection methods in \citep{gulrajani2021in}. This strategy is an oracle-selection one since we choose the model maximizing the accuracy on a validation set that follows the same distribution of the test domain.

\begin{table*}[t]
\scriptsize
\centering

\resizebox{\textwidth}{!}{
\setlength{\tabcolsep}{3pt}
\renewcommand{\arraystretch}{1.05}
\begin{tabular}{l|ccccccc|cccccc}
 & \multicolumn{7}{c|}{\textbf{RotatedMNIST}} & \multicolumn{6}{c}{\textbf{Camelyon17-WILDS}} \\

Domain & $0^\circ$ & $15^\circ$ & $30^\circ$& $45^\circ$ & $60^\circ$ & $75^\circ$ & Avg & $d_1$ & $d_2$ & $d_3$ & $d_4$ & $d_5$ & Avg\\\shline
\textsc{ERM~\citep{vapnik1998statistical}} & 96.0 ± 0.2 & 98.8 ± 0.1 & 98.8 ± 0.1 & 99.0 ± 0.0 & 99.0 ± 0.0 & 96.8 ± 0.1 & 98.1 & 96.8 ± 0.3 & 94.9 ± 0.2 & 95.9 ± 0.2 & 95.8 ± 0.2 & 94.8 ± 0.3 & 95.6 \\
\textsc{IRM~\citep{arjovsky2019invariant}} & 96.0 ± 0.2 & 98.9 ± 0.0 & 99.0 ± 0.0 & 98.8 ± 0.1 & 98.9 ± 0.1 & 95.7 ± 0.3 & 97.9 & 95.0 ± 0.7 & 92.0 ± 0.2 & 95.2 ± 0.3 & 94.3 ± 0.1 & 93.3 ± 0.6 & 94.0 \\
\textsc{GDRO~\citep{sagawa2020distributionally}} & 96.2 ± 0.1 & 98.9 ± 0.0 & 99.0 ± 0.1 & 98.7 ± 0.1 & 99.1 ± 0.0 & 96.8 ± 0.1 & 98.1 & 96.5 ± 0.1 & 95.0 ± 0.3 & 95.9 ± 0.9 & 96.0 ± 0.1 & 95.7 ± 0.4 & 95.8 \\
\textsc{Mixup~\citep{yan2020improve}} & 95.8 ± 0.3 & 98.9 ± 0.1 & 99.0 ± 0.1 & 99.0 ± 0.1 & 98.9 ± 0.1 & 96.5 ± 0.1 & 98.0 & 96.2 ± 0.0 & 94.3 ± 0.1 & 95.7 ± 0.4 & 96.7 ± 0.0 & 95.1 ± 0.1 & 95.6 \\
\textsc{MLDG~\citep{li2018learning}} & 96.2 ± 0.1 & 99.0 ± 0.0 & 99.0 ± 0.1 & 98.9 ± 0.1 & 99.0 ± 0.1 & 96.1 ± 0.2 & 98.0 & 97.0 ± 0.1 & 95.0 ± 0.3 & 96.6 ± 0.5 & 96.0 ± 0.2 & 96.1 ± 0.3 & 96.1 \\
\textsc{CORAL~\citep{sun2016deep}} & 96.4 ± 0.1 & 99.0 ± 0.0 & 99.0 ± 0.1 & 99.0 ± 0.0 & 98.9 ± 0.1 & 96.8 ± 0.2 & 98.2 & 96.5 ± 0.2 & 95.2 ± 0.1 & 96.9 ± 0.1 & 96.8 ± 0.3 & 94.8 ± 0.3 & 96.0 \\
\textsc{MMD~\citep{li2018domain}} & 95.7 ± 0.4 & 98.8 ± 0.0 & 98.9 ± 0.1 & 98.8 ± 0.1 & 99.0 ± 0.0 & 96.3 ± 0.2 & 97.9 & 96.3 ± 0.1 & 94.9 ± 0.1 & 96.8 ± 0.1 & 96.5 ± 0.2 & 93.3 ± 0.1 & 95.6 \\
\textsc{DANN~\citep{ganin2016domain}} & 96.0 ± 0.1 & 98.8 ± 0.1 & 98.6 ± 0.1 & 98.7 ± 0.1 & 98.8 ± 0.1 & 96.4 ± 0.1 & 97.9 & 93.9 ± 0.3 & 89.6 ± 1.0 & 94.5 ± 0.1 & 93.9 ± 0.5 & 92.0 ± 0.2 & 92.8 \\
\textsc{CDANN~\citep{li2018deep}} & 95.8 ± 0.2 & 98.8 ± 0.0 & 98.9 ± 0.0 & 98.6 ± 0.1 & 98.8 ± 0.1 & 96.1 ± 0.2 & 97.8 & 94.3 ± 0.1 & 91.7 ± 0.7 & 95.0 ± 0.1 & 94.7 ± 0.2 & 92.9 ± 0.5 & 93.7 \\
\rowcolor{Gray}
\textsc{\abbr} & \textbf{96.6$\pm$ 0.1} & \textbf{99.0$\pm$ 0.1} & \textbf{99.0$\pm$ 0.2} & \textbf{99.1$\pm$ 0.2} & \textbf{99.1$\pm$ 0.2} & \textbf{97.4$\pm$ 0.4} & \textbf{98.4} & \textbf{97.4$\pm$0.2} & \textbf{95.4$\pm$0.2} & 96.8$\pm$0.1 & \textbf{96.8$\pm$0.1} & \textbf{96.6$\pm$0.3} & \textbf{96.6} \\
\rowcolor{Gray}
\textsc{\abbr~w/ Aug} & \textbf{96.7$\pm$ 0.4} & \textbf{99.0$\pm$ 0.3} & \textbf{99.1$\pm$ 0.2} & \textbf{99.2$\pm$ 0.2} & 99.0$\pm$ 0.3 & \textbf{97.4$\pm$ 0.3} & \textbf{98.4} & \textbf{97.7$\pm$0.4} & \textbf{96.6$\pm$0.2} & \textbf{96.9$\pm$0.1} & \textbf{97.2$\pm$0.3} & \textbf{96.9$\pm$0.1} & \textbf{96.9} \\
\end{tabular}
}
\resizebox{\textwidth}{!}{
\setlength{\tabcolsep}{7.25pt}
\begin{tabular}{l|ccccc|ccccc}
\specialrule{0em}{8pt}{0pt}
& \multicolumn{5}{c|}{\textbf{PACS}}& \multicolumn{5}{c}{\textbf{VLCS}}\\
Domain & A & C & P & S & Avg & C & L & S & V & Avg\\\shline
\textsc{ERM~\citep{vapnik1998statistical}} & 87.8 ± 0.4 & 82.8 ± 0.5 & 97.6 ± 0.4 & 80.4 ± 0.6 & 87.2 & 97.7 ± 0.3 & 65.2 ± 0.4 & 73.2 ± 0.7 & 75.2 ± 0.4 & 77.8 \\
\textsc{IRM~\citep{arjovsky2019invariant}} & 85.7 ± 1.0 & 79.3 ± 1.1 & 97.6 ± 0.4 & 75.9 ± 1.0 & 84.6 & 97.6 ± 0.5 & 64.7 ± 1.1 & 69.7 ± 0.5 & 76.6 ± 0.7 & 77.2 \\
\textsc{GDRO~\citep{sagawa2020distributionally}} &  88.2 ± 0.7 & 82.4 ± 0.8 & 97.7 ± 0.2 & 80.6 ± 0.9 & 87.2 & 97.8 ± 0.0 & 66.4 ± 0.5 & 68.7 ± 1.2 & 76.8 ± 1.0 & 77.4 \\
\textsc{Mixup~\citep{yan2020improve}} & 87.4 ± 1.0 & 80.7 ± 1.0 & \textbf{97.9 ± 0.2} & 79.7 ± 1.0 & 86.4 & 98.3 ± 0.3 & 66.7 ± 0.5 & 73.3 ± 1.1 & 76.3 ± 0.8 & 78.7 \\
\textsc{MLDG~\citep{li2018learning}} & 87.1 ± 0.9 & 81.3 ± 1.5 & 97.6 ± 0.4 & 81.2 ± 1.0 & 86.8 & 98.4 ± 0.2 & 65.9 ± 0.5 & 70.7 ± 0.8 & 76.1 ± 0.6 & 77.8\\
\textsc{CORAL~\citep{sun2016deep}} & 87.4 ± 0.6 & 82.2 ± 0.3 & 97.6 ± 0.1 & 80.2 ± 0.4 & 86.9 & 98.1 ± 0.1 & 67.1 ± 0.8 & 70.1 ± 0.6 & 75.8 ± 0.5 & 77.8 \\
\textsc{MMD~\citep{li2018domain}} & 87.6 ± 1.2 & 83.0 ± 0.4 & 97.8 ± 0.1 & 80.1 ± 1.0 & 87.1 & 98.1 ± 0.3 & 66.2 ± 0.2 & 70.5 ± 1.0 & 77.2 ± 0.6 & 78.0\\
\textsc{DANN~\citep{ganin2016domain}} & 86.4 ± 1.4 & 80.6 ± 1.0 & 97.7 ± 0.2 & 77.1 ± 1.3 & 85.5 & 95.3 ± 1.8 & 61.3 ± 1.8 & \textbf{74.3 ± 1.0} & 79.7 ± 0.9 & 77.7 \\
\textsc{CDANN~\citep{li2018deep}} & 87.0 ± 1.2 & 80.8 ± 0.9 & 97.4 ± 0.5 & 77.6 ± 0.1 & 85.7 & 98.9 ± 0.3 & 68.8 ± 0.6 & 73.7 ± 0.6 & 79.3 ± 0.6 & 80.2 \\
\rowcolor{Gray}
\textsc{\abbr} & \textbf{88.9 ± 0.6} & \textbf{85.0 ± 1.9}  & 97.2 ± 1.2  &  \textbf{84.3 ± 0.7} & \textbf{88.9} & \textbf{99.1 ± 0.6}  & 66.5 ± 0.3  &  73.3 ± 0.6 & \textbf{80.9 ± 0.6} & 80.0 \\
\rowcolor{Gray}
\textsc{\abbr~w/ Aug} & \textbf{89.0 ± 0.3} & \textbf{86.3 ± 0.3}  & 97.0 ± 0.5  &  \textbf{84.8 ± 1.1} & \textbf{89.3} & \textbf{99.4 ± 0.2}  & \textbf{68.9 ± 2.3}  &  73.4 ± 1.1 & \textbf{81.2 ± 0.3} & \textbf{80.7} \\
\end{tabular}
}
\vspace{-0.5mm}
\caption{\footnotesize Domain generalization accuracies (\%) on RotatedMNIST, PACS, VLCS and WILDS.}
\vspace{-2mm}
\label{tab:DG_quantitative}
\end{table*}

\textbf{Model architectures.}
Following \citep{gulrajani2021in}, we use as encoders ConvNet for RotatedMNIST (detailed in Appdendix D.1 in \citep{gulrajani2021in}) and ResNet-50 for the remaining datasets.

Motivated by the observation that GAN is able to improve image quality for evaluating the disentanglement effects in the latent spaces \citep{plumerault2019controlling, shen2020interpreting, shen2020closed}, we use adversarial training \citep{goodfellow2014generative} on real samples $\mathbf{x}$ against fake ones $D(h_s(\mathbf{x}) \oplus h_v(\mathbf{\tilde{x}}))$ to attain high-quality images $\mathbf{x^{\prime}}$:
\begin{equation}
\small
    \mathcal{L}_{GAN} = \log \mathrm{Disc}(\mathbf{x})+\log(1-\mathrm{Disc}(\mathbf{x^{\prime}})).
\end{equation}
In practice, we can train the generator using adversarial training. In this stage, we employ an adversarial objective $\mathcal{L}_{GAN}$ and an additional cycle consistency constraint. With a slight abuse of notation, we denote $h_s(\cdot)$ and $h_v(\cdot)$ for $h_s(\cdot;\theta)$ and $h_v(\cdot;\phi)$, respectively. The detail of the cycle consistency constraint is that: we encode $\mathbf{x}$ and $\mathbf{\tilde{x}}$ into the latent space as $h_s(\mathbf{x}),h_s(\mathbf{\tilde{x}}), h_v(\mathbf{x}),h_s(\mathbf{\tilde{x}})$. We then swap their variation factors and generate $\mathbf{x}_{x\rightarrow \tilde{x}}=D(h_s(\mathbf{x}) \oplus h_v(\mathbf{\tilde{x}})), \mathbf{x}_{\tilde{x}\rightarrow x}=D(h_s(\mathbf{\tilde{x}}) \oplus h_v(\mathbf{x}))$. Again, the generated images will be encoded, and their variation factors will be swapped and used to generate $\mathbf{x}_{x\rightarrow \tilde{x}\rightarrow x}$ and $\mathbf{x}_{\tilde{x}\rightarrow x\rightarrow \tilde{x}}$. Finally, the cycle consistency constraint for $\mathbf{x}$ is implemented by the proposed reconstruction loss (also similarly for $\mathbf{\tilde{x}}$):
\begin{equation}
\small
\begin{aligned}
\mathcal{L}_{cyc} & = d\left( \mathbf{x}_{x\rightarrow \tilde{x} \rightarrow x}, \mathbf{x} \right) \\
& = d( D(h_s(D(h_s(\mathbf{x})) \oplus  h_v(\mathbf{\tilde{x}})))) \oplus \\ 
& \quad \quad h_v(D(h_v(\mathbf{\tilde{x}})) \oplus h_v(\mathbf{x}))))), \mathbf{x})
\end{aligned}
\end{equation}

In most experiments, the generator is a simple autoencoder, which converts the concatenation of $[h_s(\mathbf{x}),h_v(\mathbf{x})]$ to $\mathbf{x}'$. For qualitative evaluation and data augmentation experiments, the main idea of our generator follows~\citep{huang2018munit,zheng2019joint}. The decoder uses four MLPs to produce a set of AdaIN~\citep{huang2017arbitrary} parameters from the semantic factor. The variation factor is then processed by four residual blocks and four convolutional layers with these AdaIN parameters. Finally, the processed latent vector is decoded to the image space by upsampling and convolutional layers. The discriminator $D$ follows the popular multi-scale PatchGAN~\citep{isola2017image} on three input image scales: $14\times 14, 28\times 28$ and $56\times 56$. The gradient punishment~\citep{mescheder2018training} is also applied when updating $D$.

See Appendix \ref{app:exp_settings} for full details of all experimental settings including datasets statistics and visualization, baselines and its implementation, hyperparameter search and model selection protocols. See Appendix \ref{app:exp_results} for many more results.

\begin{figure}[h]
  \centering
  \includegraphics[width=\linewidth - 0.4mm]{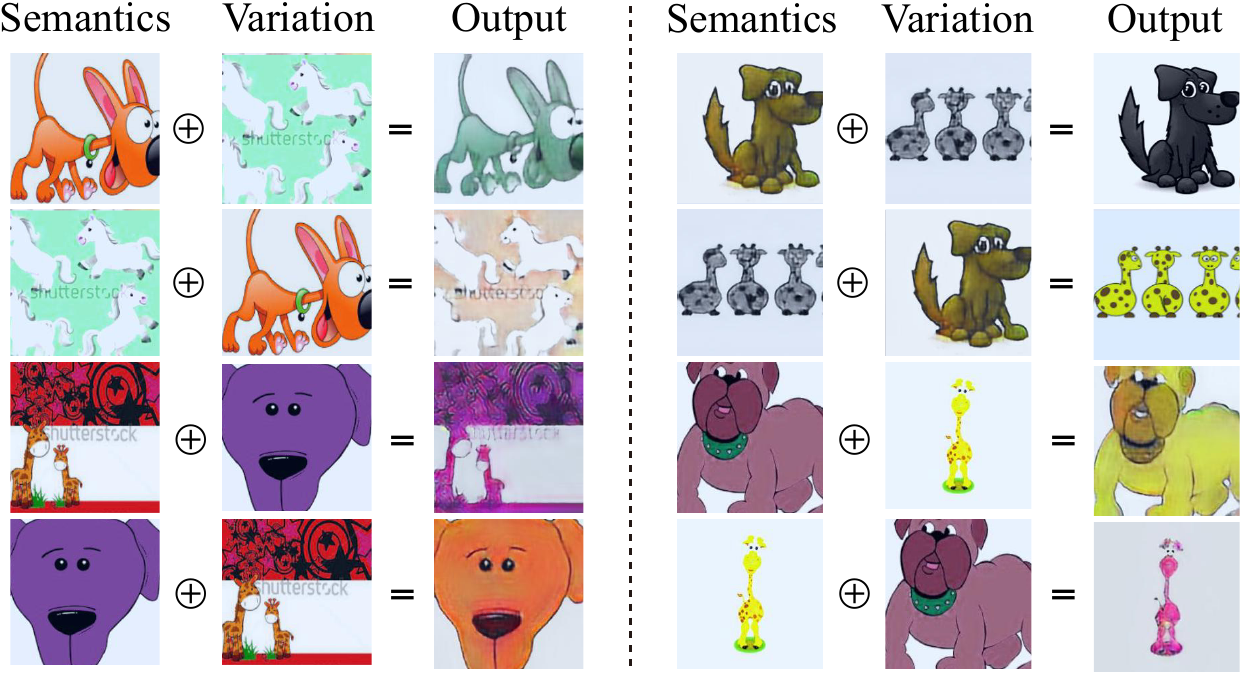}
  \caption{\footnotesize Qualitative disentanglement results. Swapping semantic and variation codes enables controllable generation for qualitative evaluation.} 
  \label{fig:main_qual}
\end{figure}

\subsection{Qualitative Studies}

We showcase some of the reconstructed images training with GAN in Fig.~\ref{fig:interpolation} and Fig.~\ref{fig:main_qual} (see appendix for many more similar results).
The results show that the representations are disentangled with respect to various variation factors like background, color etc, which supports diverse manipulations for enriching training datasets. Moreover, \abbr~captures proper semantics over data, which allows diverse manipulations on variation factors like object colors and backgrounds without changing object semantics. For example, \abbr~changes the color of a dog in the first panel of Fig. \ref{fig:main_qual} but retains the same color of its nose and eyes. Such disentanglement enables flexible and controllable generation by manipulating semantic and variation factors via swapping (Fig.~\ref{fig:main_qual}) or interpolation (Fig.~\ref{fig:interpolation}). 

\textbf{Interpolation details.} In order to better understand the learned semantic and variation representations, we further perform a linear interpolation experiment between two variation factors and generate the corresponding images as shown in Fig.~\ref{fig:interpolation}. We denote the variation code of the first and second column as $h_v(\mathbf{\tilde{x}}),h_v(\mathbf{x})$, respectively. The images from $4-13$ column are generated by $D(h_s(\mathbf{x})\oplus(i\times h_v(\mathbf{x})+(1-i)\times  h_v(\mathbf{\tilde{x}})))$ where $i\in\{1.0, 0.9, \dots, 0.1\}$. These interpolation results verify the smoothness and continuity in the variation space, and also show that our model is able to generalize in the embedding space instead of simply memorizing existing visual information. As a complementary study, we also generate images by linearly interpolating between two semantic factors while keeping the variation factors intact. We provide additional qualitative results in Appendix~\ref{app:exp_results}. 

\subsection{Numerical Results} 

Comprehensive experiments show that \abbr~consistently outperforms all the baselines by a considerable margin. From Table \ref{tab:DG_quantitative}, we observe that \abbr~achieves better DG results both in most single domains and on average. In particular, the performance gain is greater in the \textbf{worst-case} scenario like the C and S domain in PACS. This is particularly important since average performance is not an effective indicator of OOD generalization, and bad worst-case performance is tightly connected to issues like disparity amplification \citep{hashimoto2018fairness}. The performance gain of \abbr~is larger under the variation-rich dataset PACS. This makes intuitive senses because \abbr~is able to better capture inter-domain variations for improving OOD generalization.

\subsection{Empirical Analyses and Ablations}
\setlength{\columnsep}{8pt}
\begin{wrapfigure}{r}{0.242\textwidth}
  \begin{center}
  \advance\leftskip+1mm
  \renewcommand{\captionlabelfont}{\footnotesize}
    \vspace{-0.3in}  
    \includegraphics[width=0.242\textwidth]{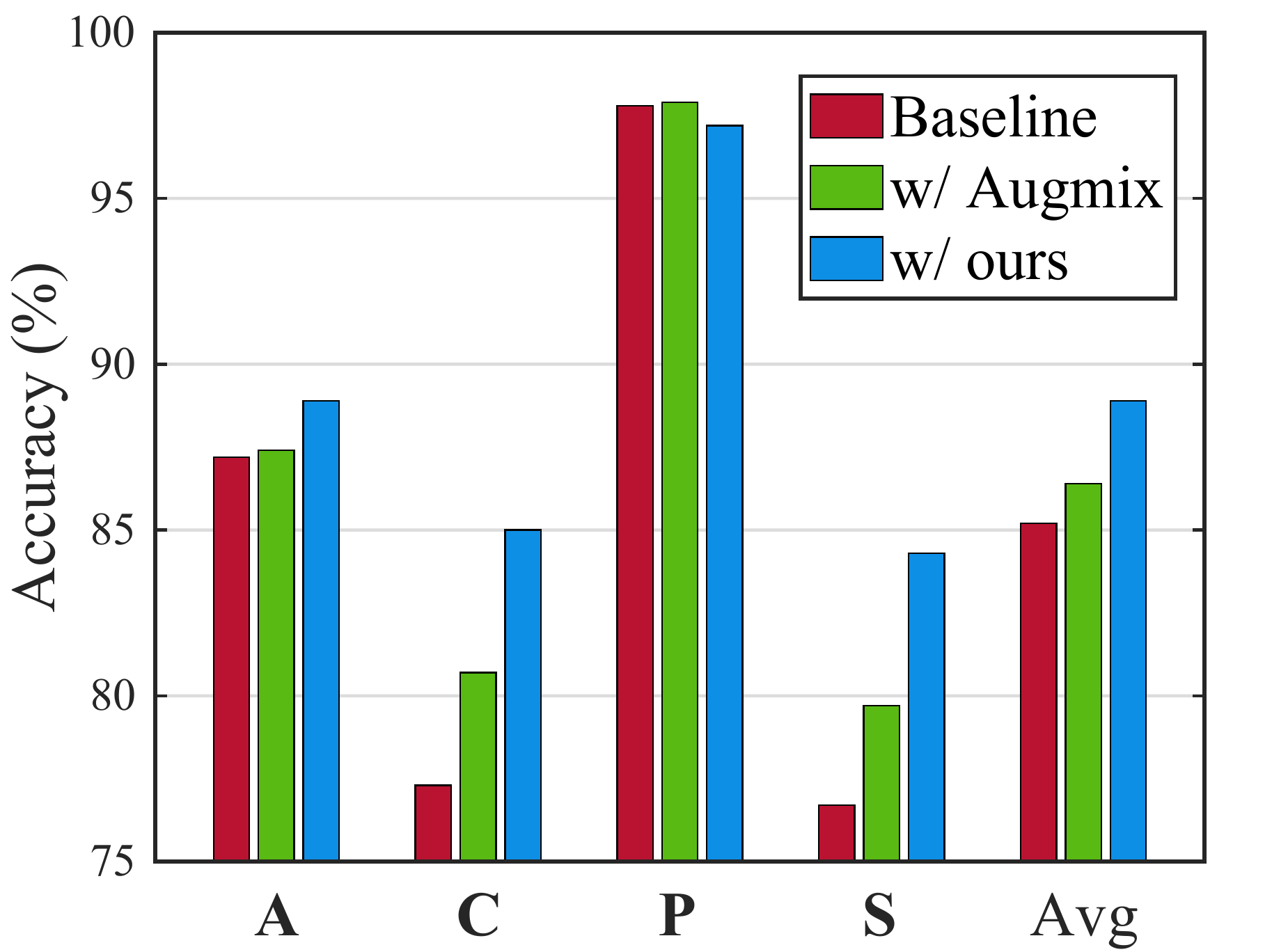}
    \vspace{-0.26in} 
    \caption{\footnotesize Data augmentation on PACS with different target domain.}
  \label{fig:abl_pacs}
    \vspace{-0.2in} 
  \end{center}
\end{wrapfigure}

\textbf{Effect of data augmentation.} We first evaluate the effect of data augmentation by comparing our learned data augmentation network with a heuristic-based augmentation method AugMix \citep{hendrycks*2020augmix}. Fig.~\ref{fig:abl_pacs} shows that the constraints optimization brings great performance gain over vanilla ERM and a data augmentation heuristics Augmix, especially the worst-case (i.e. the S domain in PACS) performance. The effectiveness of the data augmentation procedure in \abbr~is well connected to many empirical evidence in \citep{hendrycks2020pretrained, tu2020empirical, zhang2019unseen} and also validates the hypothesis in \citep{montero2021the} that the richness of training domain data is crucial for extrapolation.

\setlength{\columnsep}{8pt}
\begin{wrapfigure}{r}{0.242\textwidth}
  \begin{center}
  \advance\leftskip+1mm
  \renewcommand{\captionlabelfont}{\footnotesize}
    \vspace{-0.17in}  
    \includegraphics[width=0.242\textwidth]{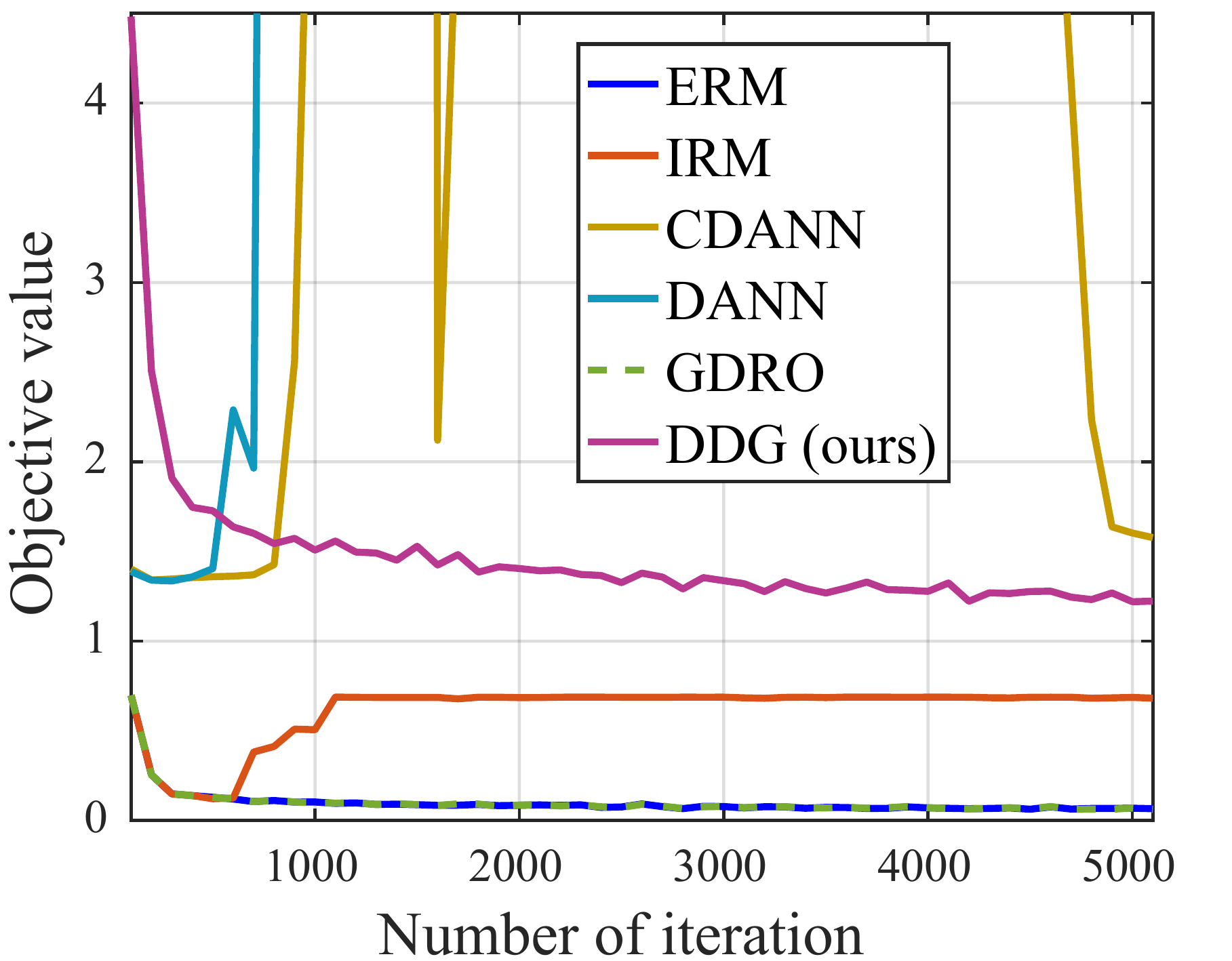}
    \vspace{-0.26in} 
    \caption{\footnotesize Convergence comparison.}
        \label{fig:converge_exp}
    \vspace{-0.2in} 
  \end{center}
\end{wrapfigure}

\textbf{Convergence analysis.} We investigate the training dynamics of \abbr~and several baselines over WILDS, where the target domain is $d_5$. The learning curves in Fig.~\ref{fig:converge_exp} show that domain adversarial training methods like DANN, CDANN are unstable and hard to converge due to their adversarial training nature. IRM has a similar pattern yet is more stable. Thanks to the primal-dual algorithm, \abbr~can converge much better than the above methods, bear a resemblance to the ERM counterpart. 

\setlength{\columnsep}{8pt}
\begin{wrapfigure}{r}{0.242\textwidth}
  \begin{center}
  \advance\leftskip+1mm
  \renewcommand{\captionlabelfont}{\footnotesize}
    \vspace{-0.22in}  
    \includegraphics[width=0.242\textwidth]{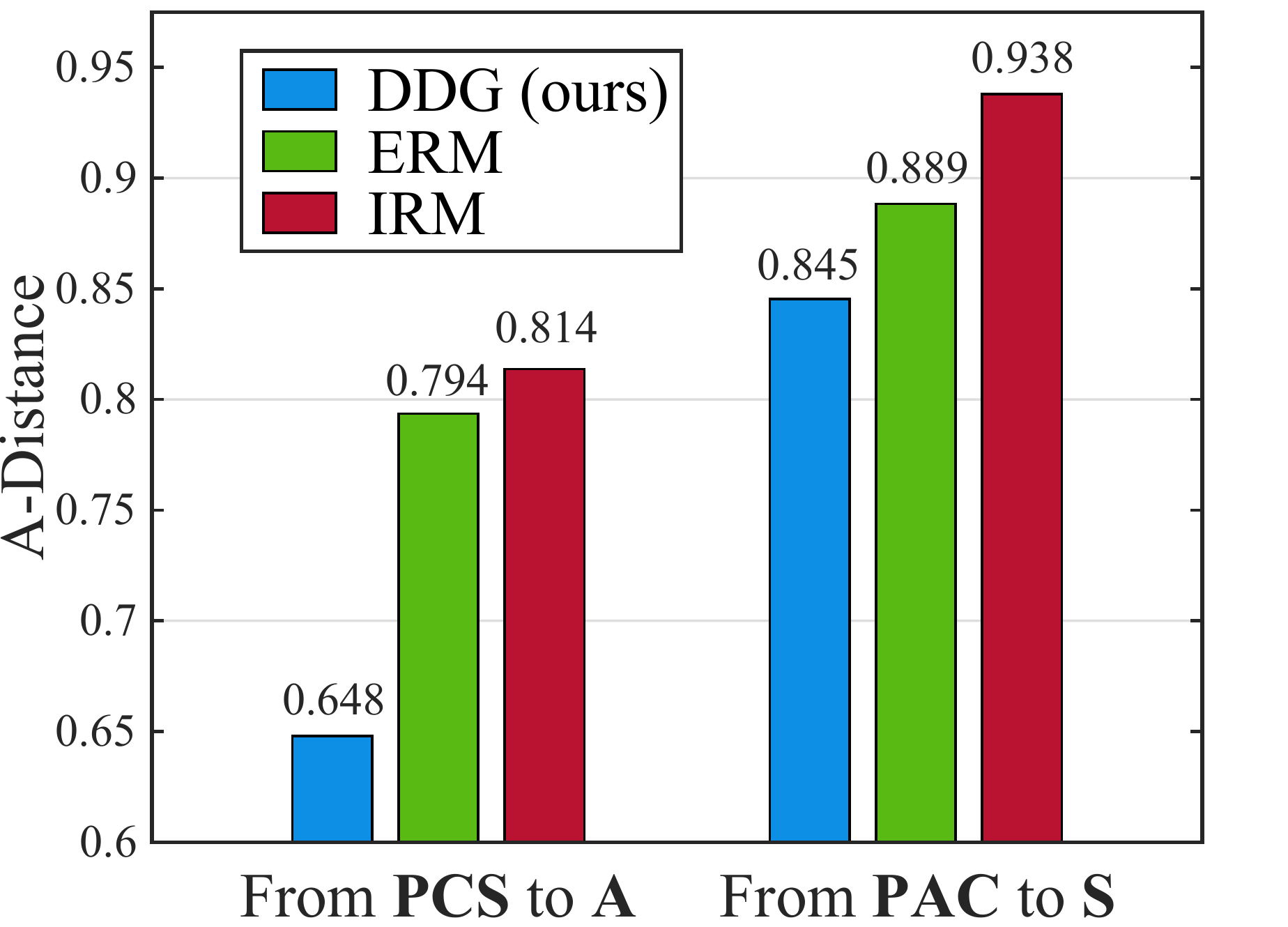}
    \vspace{-0.23in} 
    \caption{\footnotesize $\mathcal{A}$-distance on learned features for different generalization tasks.}\label{fig:a-dist_exp}
    \vspace{-0.17in} 
  \end{center}
\end{wrapfigure}

\textbf{Evaluation of domain divergence.} We use the $\mathcal{A}$-distance to measure domain discrepancy \citep{ben2010theory}. This can be approximated as $\thickmuskip=2mu \medmuskip=2mu d_{\mathcal{A}} = 2(1 - 2\sigma)$, where $\sigma$ is the error of a two-sample classifier distinguishing features of samples from source and target domains \citep{long2015learning}.
Fig. \ref{fig:a-dist_exp} shows that \abbr~can learn more invariant features to \textit{minimize the divergence} between source and target domains than IRM and ERM.

\textbf{Qualitative comparison with AugMix.} Both Fig.~\ref{fig:main_qual} and Fig.~\ref{fig:add_pacs} in Appendix~\ref{app:exp_results} show that \abbr~produces samples with diverse styles. In contrast, Fig.~\ref{fig:augmix} in Appendix~\ref{app:exp_results} shows that it is much more difficult for heuristic-based methods such as AugMix to generate samples with diverse styles for training. The qualitative results validate the effectiveness of \abbr~as an automatic data augmentation method.

%% file: 6-conclusions.tex
\section{Concluding Remarks}\label{sec:conclusion}

We propose a novel disentangled learning framework for domain generalization, with both theoretical analyses and practical algorithmic implementation. By separating semantic and variation representations into different subspaces while enforcing invariance constraints, \abbr~yields superior OOD performance with improved empirical convergence and also yields interpretable and controllable generative results. In this work, we only consider the disentangled effects between semantic and variation factors since it is hard to provide known generative variation factors that manifest the distribution shifts precisely. It remains an open problem to improve the disentanglement between different variation factors with limited supervision and evaluate the treatment effects of data augmentation in a controllable manner~\cite{zhang2022can}.

\newpage

%% file: 7-appendix.tex
\newpage
\onecolumn
\begin{appendix}
\begin{center}
{\LARGE \textbf{Appendix}}
\end{center}

\section{Proofs}
\subsection{Proof for parameterization gap}
\begin{proof}
Recall the feasibility assumption \ref{assump:feasibility} ensures the strong duality such that the primal and dual optimal objectives are equal, which means for $\forall$ $\tilde{\lambda} \in \mathbb{R}^+$, $\tilde{f}_s, \tilde{f}_v \in \mathcal{H}$, the following saddle point condition holds 

\begin{equation}
\begin{aligned}
    \mathcal{R}(f_s^*,f_v^*,\tilde{\lambda}) \le &\max\limits_{\lambda} \min\limits_{f_s, f_v} \mathcal{R}(f_s,f_v,\lambda) = D^* = P^* \\
    = & \min_{f_s, f_v} \max_{\lambda} \mathcal{R}(f_s, f_v, \lambda) \le \mathcal{R}(\tilde{f}_s,\tilde{f}_v,\lambda^*)
    \label{eq:saddle_1}
\end{aligned}
\end{equation}

Recall the definition of $\DD_{\epsilon}^*$ and by the inclusion relation $\mathcal{H}_\theta \subseteq \mathcal{F}$, we can derive the lower bound as
\begin{equation}
\begin{aligned}
    \DD_{\epsilon}^{*}(\gamma) & \triangleq \max _{\lambda} \min _{\theta, \phi \in \mathcal{H}} \mathcal{L}(\theta)+\lambda \mathcal{L}_{\operatorname{con}}(\theta, \phi) \\
    & \ge \min _{\theta, \phi \in \mathcal{H}} \mathcal{L}(\theta)+\lambda \mathcal{L}_{\operatorname{con}}(\theta, \phi), \forall \lambda \in \mathbb{R}^+ \\
    & \ge \min_{f_s, f_v \in \mathcal{F}} \mathcal{L}(f_s)+\lambda \mathcal{L}_{\operatorname{con}}(f_s, f_v) = \PP^*
\end{aligned}
\end{equation}

For the upper bound, we add and subtract $\mathcal{R} \equiv \mathcal{R}(f_s, f_v, \lambda) = \min\limits_{f_s\in\mathcal{F}} \mathcal{L}(f_s) + \min\limits_{f_s, f_v\in\mathcal{F}} \lambda\mathcal{L}_{con}(f_s, f_v)$ from $\DD^{\star}_{\epsilon}(\gamma)$ to get
\begin{equation}
\begin{aligned}
    \DD^{\star}_{\epsilon}(\gamma) &= \max\limits_{\lambda} \min\limits_{\theta,f_s,f_v} \mathcal{L}(\theta) + \lambda \mathcal{L}_{con}(\theta, \phi) +  \mathcal{R} - \mathcal{R} \\
    &= \max\limits_{\lambda} \min\limits_{\theta,f_s,f_v} \mathcal{R} +  [\mathcal{L}(\theta)-\mathcal{L}(f_s)] + \lambda [\mathcal{L}_{con}(\theta, \phi) - \mathcal{L}_{con}(f_s, f_v)] \\
    \label{eq:inter_param}
\end{aligned}
\end{equation}

With a slight abuse of notation, $\mathbb{E}$ is short for $\mathbb{E}_{(\x,y) \sim \mathbb{P}(X,Y),\xt \sim \mathbb{P}(X)}$. 
Then we consider the combination of the second and third term as 
\begin{equation}
\begin{aligned}
    & [1, \lambda] 
    \left[\begin{array}{c}
\mathcal{L}(\theta)-\mathcal{L}\left(f_{s}\right)  \\
\mathcal{L}_{\text {con }}(\theta, \phi)-\mathcal{L}_{\text {con }}\left(f_{s}, f_{v}\right) \\
\end{array}\right] \\
     \le& (1+ \| \lambda \|_1) \left\| \left[\begin{array}{c}
\mathcal{L}(\theta)-\mathcal{L}\left(f_{s}\right)  \\
\mathcal{L}_{\text {con }}(\theta, \phi)-\mathcal{L}_{\text {con }}\left(f_{s}, f_{v}\right) \\
\end{array}\right] \right\|_{\infty}
\\
     =& (1+|\lambda|) \max \left\{ \mathcal{L}(\theta)-\mathcal{L}\left(f_{s}\right), \mathcal{L}_{\text {con }}(\theta, \phi)-\mathcal{L}_{\text {con }}\left(f_{s}, f_{v}\right) \right\} \\
     =& (1+ |\lambda|) \max \left\{ \mathbb{E}[\ell\left(h_{s}(\mathbf{x} ; \theta) ; y\right)-\ell\left(f_{s}(\mathbf{x}) ; y\right)],
    \mathbb{E}[d\left(\mathbf{x}, D\left(h_{s}(\mathbf{x} ; \theta), 
    h_{v}(\tilde{\mathbf{x}} ; \phi)\right)\right)-d\left(\mathbf{x}, D\left(f_{s}(\mathbf{x}), f_{v}(\tilde{\mathbf{x}})\right)\right)] \right\} \\
    \le & (1+|\lambda|) \mathbb{E} \left\{ \max \left[ [\ell\left(h_{s}(\mathbf{x} ; \theta) ; y\right)-\ell\left(f_{s}(\mathbf{x}) ; y\right)], \mathbb{E}\left[d\left(\mathbf{x}, D\left(h_{s}(\mathbf{x} ; \theta), 
    h_{v}(\tilde{\mathbf{x}} ; \phi)\right)\right)-d\left(\mathbf{x}, D\left(f_{s}(\mathbf{x}), f_{v}(\tilde{\mathbf{x}})\right)\right)\right] \right] \right\} \\
    \le & (1+|\lambda|) \mathbb{E} \{ \max [ L_{\ell} \left|h_{s}(\mathbf{x} ; \theta)-f_{s}(\mathbf{x})\right|,  L_d |D\left(h_{s}(\mathbf{x} ; \theta), h_{v}(\tilde{\mathbf{x}} ; \phi)\right)-D\left(f_{s}(\mathbf{x}), f_{v}(\tilde{\mathbf{x}})\right)| ]\} \\
    \le & (1+|\lambda|) \mathbb{E} \max \{ L_\ell \epsilon_s, L_d \epsilon_g \} \\
    =& (1+|\lambda|) \max \{ L_\ell \epsilon_s, L_d \epsilon_g \}
\end{aligned}
\end{equation}

where the first inequality is using Hölder's inequality (\ref{eq:holder}) when $p=1, q=\infty$ and the second one is by the convexity of max-norm and Jensen's inequality. The third inequality is applying $L_\ell$ and $L_d$ lipschitzness on $\ell$ and $d$. The fourth one is due to $\epsilon_s$ and $\epsilon_g$ parameterization on $f_s$ and $D$.
\begin{equation}
    \sum_{k=1}^{n}\left|x_{k} y_{k}\right| \leq\left(\sum_{k=1}^{n}\left|x_{k}\right|^{p}\right)^{1 / p}\left(\sum_{k=1}^{n}\left|y_{k}\right|^{q}\right)^{1 / q}
    \label{eq:holder}
\end{equation}

Then Eq. (\ref{eq:inter_param}) becomes
\begin{equation}
\begin{aligned}
    \DD_{\epsilon}^{\star}(\gamma) & \le \max _{\lambda} \min _{f_{x}, f_{v}} 
    \mathcal{R} + (1+|\lambda|) \max \{ L_\ell \epsilon_s, L_d \epsilon_g \}  \triangleq D_p^*
\end{aligned}
\end{equation}

In order to use strong duality to bound $\DD_p^*$, we need to construct primal problem whose optimum is $\PP_p^*$ from $\DD_p^*$. Therefore, the remaining proof is to show $\DD_p^*$ is the dual problem to a constraint statistical learning problem with perturbation function as $m = \max \left\{L_{\ell} \epsilon_{s}, L_{d} \epsilon_{g}\right\}$. Specifically, when $\lambda > 0$, $D_p^*$ can be expanded and rearranged as 
\begin{equation}
\begin{aligned}
    \DD_p^* & = \max_{\lambda} \min_{f_s, f_v} \mathcal{L}\left(f_{s}\right)+ \lambda \mathcal{L}_{\operatorname{con}}\left(f_{s}, f_{v}\right) +(1+|\lambda|) m \\
    & = \max_{\lambda} \min_{f_s, f_v} \ell\left(f_{s}(\mathbf{x} ; \theta) ; y\right) + m +  \lambda [d\left(\mathbf{x}, D\left(f_{s}(\mathbf{x}; \theta), f_{v}(\tilde{\mathbf{x}} ; \phi)\right)\right) - \gamma + m] \\
\label{eq:rearange}
\end{aligned}
\end{equation} 
By the feasibility assumption \ref{assump:feasibility}, Eq. (\ref{eq:rearange}) can be considered as the dual problem of the following one: 
\begin{equation}
\begin{aligned}
&\PP^{\star}_p(\gamma) \triangleq \min _{f_{s} \in \mathcal{F}} \mathcal{L}\left(f_{s}\right) +m \\
&\text{s.t.} \quad d\left(\mathbf{x}, D\left(f_{s}(\mathbf{x} ; \theta), f_{v}(\tilde{\mathbf{x}} ; \phi)\right)\right) \leq \gamma -m
\end{aligned}
\end{equation}

Denote $f_s^*$ and $f_v^*$ as the primal solutions to $\PP_p^*$, $\lambda^*$ and $\lambda_p^*$ as the dual variable that give solutions $\DD_p^*$, $\PP_p^*$. With the regularity assumption \ref{assump:regularity}, the saddle point condition shows for $\forall \tilde{f}_s, \tilde{f}_v \in \mathcal{H}$ and $\lambda, \lambda_p^* \in \mathbb{R}^+$, it holds that
\begin{equation}
\begin{aligned}
     \mathcal{R}(f_{s,p}^*, f_{v,p}^*, \tilde{\lambda}) + (1+|\tilde{\lambda}|) & \le \max_\lambda \min_{f_s, f_v} \mathcal{R}(f_s, f_v, \lambda) + (1+|\lambda|)m = \DD_p^* = \PP_p^* \\
    & = \min_{f_s, f_v} \max_\lambda \mathcal{R}(f_s, f_v, \lambda) + (1+|\lambda|)m \\
    & \le \mathcal{R}(\tilde{f_s}, \tilde{f_v}, \lambda_p^*) + (1+|\lambda_p^*|)m
\label{eq:saddle_R}
\end{aligned}
\end{equation}

Recall that Eq. (\ref{eq:saddle_R}) holds for $\forall \tilde{f}_s, \tilde{f}_v \in \mathcal{H}$. Let $\tilde{f}_s=f^*_s, \tilde{f}_v=f^*_v$, then we use it to upper bound $D_\epsilon^*$ as
\begin{equation}
\begin{aligned}
    \DD_\epsilon^*(\gamma) \le \PP_p^* & \le \mathcal{R}(f^*_s, f^*_v, \lambda^*) + (1+|\lambda_p^*|)m  = \PP^*(\gamma) + (1+|\lambda_p^*|)m
\end{aligned}
\end{equation}
The last step is due to Eq. (\ref{eq:saddle_1}).
\label{app:param_proof}
\end{proof}

\subsection{Proof for empirical gap}
\begin{proof}
Similar to \citep{chamon2020empirical, robey2021model}, by KKT conditions and complementary slackness conditions \citep{boyd2004convex} shows

\begin{equation}
\begin{aligned}
    & \lambda^*_\epsilon (\mathbb{E}_{\x,\xt\sim\mathbb{P}(X)} \left[d\left(\x, D\left(f_{s}(\x ; \theta^*_\epsilon), f_{v}(\xt ; \phi^*_\epsilon)\right)\right)-\gamma \right] = 0 \\
    & \lambda^*_{\epsilon, n} (\sum\limits_{i=1}^n \sum\limits_{j\neq i}^n d\left(\mathbf{x_i}, D\left(f_{s}(\mathbf{x_i} ; \theta^*_\epsilon), f_{v}(\mathbf{x_j} ; \phi^*_\epsilon)\right) -\gamma \right) = 0 
\label{eq:KKT}
\end{aligned}
\end{equation}

where $(\theta^*_\epsilon, \phi^*_\epsilon, \lambda^*_\epsilon)$ and $(\theta^*_{\epsilon,n}, \phi^*_{\epsilon,n}, \lambda^*_{\epsilon,n})$ are primal-dual pairs for achieving the optimum $\DD_\epsilon^*(\gamma)$ and $\DD_{\epsilon,n}^*(\gamma)$.

Eq. (\ref{eq:KKT}) implies the constraint-related terms in the objectives to be zero, then consider the remaining term as 
\begin{equation}
\begin{aligned}
    & \DD_\epsilon^*(\gamma) = \mathbb{E}\left[\ell\left(f_{s}(\mathbf{x} ; \theta) ; y\right)\right] \triangleq \MM(\theta^*_\epsilon)\\
    & \DD_{\epsilon,n}^*(\gamma) = \sum_{i=1}^{n} \ell\left(f_{s}\left(\mathbf{x}_{\mathbf{i}}\right), y_{i}\right) \triangleq \hat{M}(\theta^*_{\epsilon,n})
\end{aligned}
\end{equation}

Thus the empirical gap reduces to
\begin{equation}
|\DD_\epsilon^*(\gamma) - \DD_{\epsilon,n}^*(\gamma)| = |\MM(\theta^*_\epsilon) - \hat{\MM}(\theta^*_{\epsilon,n})|
\label{eq:reduced_gap}
\end{equation}

Using the fact that $\theta^*_\epsilon$ and $\theta^*_{\epsilon,n}$ are optimal for $\MM(\theta^*_\epsilon)$ and $\hat{\MM}(\theta^*_{\epsilon,n})$, the following holds
\begin{equation}
\begin{aligned}
    \MM(\theta^*_\epsilon) - \hat{\MM}(\theta^*_{\epsilon}) & \le \MM(\theta^*_\epsilon) - \hat{\MM}(\theta^*_{\epsilon,n})  \le  \MM(\theta^*_{\epsilon,n}) - \hat{\MM}(\theta^*_{\epsilon,n})
\end{aligned}
\end{equation}
Therefore, using the above lower and upper bound, we can bound Eq. (\ref{eq:reduced_gap}) as 
\begin{equation}
\begin{aligned}
    |\DD_\epsilon^*(\gamma) - \DD_{\epsilon,n}^*(\gamma)| \le \max \{ & |\MM(\theta^*_\epsilon) - \hat{\MM}(\theta^*_{\epsilon})|, |\MM(\theta^*_{\epsilon,n}) - \hat{\MM}(\theta^*_{\epsilon,n})| \}
\end{aligned}
\label{eq:max_m_diff}
\end{equation}
Then we resort to the classical VC-dimension bounds for the above two terms in Eq. (\ref{eq:max_m_diff}) as follows
\begin{equation}
    |\MM(\theta) - \hat{\MM}(\theta)| \le 2 B \sqrt{\frac{1}{n}\left[1+\log \left(\frac{4(2 n)^{d_{\mathrm{VC}}}}{\delta}\right)\right]}
    \label{eq:VC_final}
\end{equation}
holds with probability $1-\delta$ when $d_{VC}$ is the VC dimension for all $\theta$.

Combing Eq. (\ref{eq:max_m_diff}) and (\ref{eq:VC_final}) completes the proof.
\label{app:emp_proof}
\end{proof}

\subsection{Proof for empirical duality gap}

\begin{proof}
Simply combining the results in the above lemmas, i.e. parameterization gap and empirical gap, via applying the triangle inequality completes the proof.
\begin{equation*}
\small
\begin{aligned}
 \left|\PP^{\star}-\DD_{\varepsilon, n}^{\star}(\gamma)\right|=& \left|\PP^{\star}+\DD_{\epsilon}^{\star}(\gamma)-\DD_{\epsilon}^{\star}(\gamma)-\DD_{\epsilon, n}^{\star}(\gamma)\right|\\
 \leq & |\PP^{\star}-\DD_{\varepsilon}^{\star}(\gamma)|+ |\DD_{\epsilon}^{\star}(\gamma)-\DD_{\epsilon, n}^{\star}(\gamma)|\\
 \leq & (1+|\lambda|) m + 2B \sqrt{\frac{1}{n}\left[1+\log \left(\frac{4(2 n)^{d_{v c}}}{\delta}\right)\right]}
\end{aligned}
\end{equation*}
\label{app:emp_duality_proof}
\end{proof}

\newpage

\section{Domain Generalization by Learning on Fictitious Distributions}
\label{app:da_bound}
This section gives a justification for disentanglement from a different perspective by connecting the dots with classical domain adaptation. Specifically, we construct a fictitious distribution to extend it to the DG setting and decompose the target learning objective into empirical learning errors, domain divergence and source domain data diversity. Moreover, we show that learning disentangled representations gives a tighter risk upper bound.

With a slight abuse of notation, let $\mathcal{H}$ be a hypothesis space and denote $\mathcal{\tilde{D}}$ as the induced distribution over feature space $\mathcal{Z}$ for every distribution $\mathcal{D}$ over the raw space. Define $\mathcal{D}_{S}^{i}$ as the source distribution over $\mathcal{X}$, which enables a mixture construction of source domains as $\mathcal{D}_S^\alpha=\sum_{i=1}^{N_s}\alpha_i\mathcal{D}_S^i(\cdot)$. Denote a fictitious distribution $\mathcal{D}_U^\alpha=\sum_{i=1}^{N_s}\alpha^*_i\mathcal{D}_S^i(\cdot)$ as the convex combination of source domains which is the closest to $\mathcal{D}_U$, where $\alpha^*_1,...,\alpha^*_{N_S}=\arg\min_{\alpha_1,...,\alpha_{N_s}}d_{\mathcal{H}}(\mathcal{D}_U,\sum_{i=1}^{N_s}\alpha_i\mathcal{D}_S^i(\cdot))$. The fictitious distribution induces a feature space distribution $\tilde{\mathcal{D}}_U^\alpha=\sum_{i=1}^{N_s}\alpha^*_i\tilde{\mathcal{D}}_S^i(\cdot)$. The following inequality holds for the risk $\epsilon_{U}(h)$ on any unseen target domain $\mathcal{D}_U$. 
\begin{equation}
    \epsilon_U(h) \leq \lambda_\alpha+\underbrace{\sum_{i=1}^{N_S}\alpha_i\epsilon_{S,i}(h)}_{\rm \circled{\rm 1} Empirical}+\underbrace{d_{\mathcal{H}}(\tilde{\mathcal{D}}^\alpha_U,\tilde{\mathcal{D}}_S^\alpha)}_{\rm \circled{\rm 2} Divergence}+\underbrace{d_{\mathcal{H}}(\tilde{\mathcal{D}}_U,\tilde{\mathcal{D}}^\alpha_U)}_{\rm \circled{\rm 3} Diversity}
    \label{lemma:bound}
\end{equation}
where $\lambda_\alpha$ is the risk of the optimal hypothesis on the mixture source domain $\mathcal{D}_S^\alpha$ and $\mathcal{D}_U$.

We define the symmetric difference hypothesis space as $\mathcal{H} \Delta \mathcal{H}=\left\{h(\mathbf{x}) \oplus h^{\prime}(\mathbf{x}): h, h^{\prime} \in \mathcal{H}\right\}$, where $\oplus$ is the XOR operator. Applying~\citep{2006Analysis}, we have
\begin{equation}
\begin{aligned}
    \epsilon_U(h)&\leq\lambda_U+\text{Pr}_{\mathcal{D}_U}[\mathcal{Z}_h\triangle\mathcal{Z}_h^*] \\ & \leq  \lambda_U+\text{Pr}_{\mathcal{D}_{S}^\alpha}[\mathcal{Z}_h\triangle\mathcal{Z}_h^*]+|\text{Pr}_{\mathcal{D}_{S}^\alpha}[\mathcal{Z}_h\triangle\mathcal{Z}_h^*]-\text{Pr}_{\mathcal{D}_U}[\mathcal{Z}_h\triangle\mathcal{Z}_h^*]|\\
    & \leq \lambda_U+\text{Pr}_{\mathcal{D}_{S}^\alpha}[\mathcal{Z}_h\triangle\mathcal{Z}_h^*]+d_{\mathcal{H}}(\tilde{D}_U,\tilde{\mathcal{D}}_S^\alpha)\\
    & \leq \lambda_U+\text{Pr}_{\mathcal{D}_{S}^\alpha}[\mathcal{Z}_h\triangle\mathcal{Z}_h^*]+d_{\mathcal{H}}(\tilde{\mathcal{D}}^\alpha_U,\tilde{\mathcal{D}}_S^\alpha)+d_{\mathcal{H}}(\tilde{\mathcal{D}}_U,\tilde{\mathcal{D}}^\alpha_U)\\
    & \leq \lambda_\alpha+\underbrace{\sum_{i=1}^{N_S}\alpha_i\epsilon_{S,i}(h)}_{\rm \circled{\rm 1} Empirical}+\underbrace{d_{\mathcal{H}}(\tilde{\mathcal{D}}^\alpha_U,\tilde{\mathcal{D}}_S^\alpha)}_{\rm \circled{\rm 2} Divergence}+\underbrace{d_{\mathcal{H}}(\tilde{\mathcal{D}}_U,\tilde{\mathcal{D}}^\alpha_U)}_{\rm \circled{\rm 3} Diversity}
\end{aligned}
\label{proof:boundlabel}
\end{equation}

The fourth inequality holds because of the triangle inequality. We provide the explanation for our bound in the Eq. (\ref{lemma:bound}) and Eq. (\ref{proof:boundlabel}). The second term is the empirical loss for the convex combination of all source domains. The third term corresponds to ``To what extent can the convex combination of the source domain approximate the target domain''. The minimization of the third term requires diverse data or strong data augmentation, such that the unseen distribution lies within the convex combination of source domains. For the fourth term,the following equation holds for any two distributions $D_U',D_U''$, which are the convex combinations of source domains \citep{carlucci2019domain}
\begin{equation}
    \begin{aligned}
        d_{\mathcal{H}[D_U',D_U'']}\leq \sum_{l=1}^{N_S}\sum_{k=1}^{N_S} \alpha_l\alpha_kd_{\mathcal{H}}[\mathcal{D}_{S,l},\mathcal{D}_{S,k}]
    \end{aligned}
\end{equation}
Such an upper bound will be minimized when $d_{\mathcal{H}}[\mathcal{D}_{S,l},\mathcal{D}_{S,k}]=0,\forall \, l,k\in\{1,...,N_S\}$. Namely projecting the source domain data into a feature space, where the source domain labels are hard to distinguish.

The above $\rm \circled{\rm 3}$ Diversity term is also supported by the evidence that compositional generalization and extrapolation can be improved if the training domain data are rich enough \citep{chaabouni2020compositionality, montero2021the, Hill2020Environmental}. To this end, one can obviously simulate data points with predetermined data augmentation methods such as rotating, cropping, Gaussian blur, color jitter, etc.
However, their developments require prior knowledge and domain-specific expertise like translation-invariance on images, which is likely to fail in the unseen domain due to distribution shifts. It motivates learning disentangled representations that are transferable across various domains \citep{dittadi2021on}. Thus we discuss the benefits of disentanglement on the domain generalization gap in the following section. 
Assume that the semantic and the variation factors are disentangled in the latent space $\mathcal{S}$ and $\mathcal{V}$, then the errors~\citep{Ruichu2019Learning} on the disentangled source and target domain with a hypothesis $h$ are
 \begin{equation}
     \epsilon_{S,i}(h)=\epsilon_{S,i}^s(h)+\epsilon_{S,i}^v(h),  \epsilon_U(h)=\epsilon_{U}^s(h)+\epsilon_{U}^v(h)
     \label{lemma:disentangledbound}
 \end{equation}

Given $h^* = \arg\min_{h \in \mathcal{H}}\left(\epsilon_{S,i}^s(h), \epsilon_{S,i}^v(h)\right)$ $ \forall i  \in \{1, \dots, N_S\}$, since $\epsilon_U(h)=\epsilon_{U}^s(h)+\epsilon_{U}^v(h)$, combining Eq. (\ref{lemma:bound}) and we have
\begin{equation}
\epsilon^s_U(h) + \epsilon_{U}^v(h) \leq \lambda_\alpha+\sum_{i=1}^{N_S}\alpha_i\epsilon_{S,i}(h)+d_{\mathcal{H}}(\tilde{\mathcal{D}}^\alpha_U,\tilde{\mathcal{D}}_S^\alpha)+d_{\mathcal{H}}(\tilde{\mathcal{D}}_U,\tilde{\mathcal{D}}^\alpha_U)
\end{equation}

where $\lambda_\alpha=\epsilon_{U}(h^*)+\sum_{i=1}^{N_S}\alpha_i\epsilon_{S,i}(h^*)=\epsilon^s_{U}(h^*)+\epsilon^v_{U}(h^*)+\sum_{i=1}^{N_S}\alpha_i\epsilon_{S,i}^s(h^*)+\sum_{i=1}^{N_S}\alpha_i\epsilon^v_{S,i}(h^*)$. Then the upper bound for the unseen domain can be further derived as follow,

\begin{equation}
    \epsilon^s_U(h) \leq \lambda_\alpha+ \sum_{i=1}^{N_S}\alpha_i\epsilon_{S,i}(h)+d_{\mathcal{H}}(\tilde{\mathcal{D}}^\alpha_U,\tilde{\mathcal{D}}_S^\alpha)+d_{\mathcal{H}}(\tilde{\mathcal{D}}_U,\tilde{\mathcal{D}}^\alpha_U) - \epsilon_{U}^v(h)
\end{equation}

Combining Eq.~(\ref{lemma:bound}) and Eq.~(\ref{lemma:disentangledbound}), we have
\begin{equation}
    \epsilon_U^s(h) \leq
    \lambda_\alpha+\underbrace{\sum_{i=1}^{N_S}\alpha_i\epsilon_{S,i}(h)}_{\rm \circled{\rm 1} Empirical}+\underbrace{d_{\mathcal{H}}(\tilde{\mathcal{D}}^\alpha_U,\tilde{\mathcal{D}}_S^\alpha)}_{\rm \circled{\rm 2} Divergence}+\underbrace{d_{\mathcal{H}}(\tilde{\mathcal{D}}_U,\tilde{\mathcal{D}}^\alpha_U)}_{\rm \circled{\rm 3} Diversity} - \underbrace{\epsilon_U^v(h)}_{\rm \circled{\rm 4}} 
    \label{lemma:disen_bound}
\end{equation}

where $\rm \circled{\rm 1}$denotes the empirical loss over every source domain,$\rm \circled{\rm 2}$means divergence minimization among source domains and$\rm \circled{\rm 3}$encourages the diversity and coverage of the mixture of source domains. The above result shows the benefits of disentanglement over representation spaces. Thus we propose to use two separate encoders representing semantic and variation subspaces respectively in the following section. Such a formulation combined with the disentanglement term $\rm \circled{\rm 4}$ in Eq. (\ref{lemma:disen_bound}) implies that we should perform ERM over the semantic space only. 
The above analysis essentially justifies the design of \abbr~from a classical domain adaptation perspective: optimizing empirical risk over semantic space while promoting diversity and divergence by taking disentanglement as a constraint.

\newpage

\section{Experimental Settings}
\label{app:exp_settings}

\subsection{Other Training Details}
We optimize all models using Adam \citep{kingma2015adam}. For all the detailed hyperparameter settings, please refer to our code which is publicly available at \hyperlink{https://github.com/hlzhang109/DDG}{https://github.com/hlzhang109/DDG}.

\subsection{Dataset Statistics and Visualization}
\label{app:dataset}
\begin{figure*}[h]
    \centering
     \begin{subfigure}[t]{.33\linewidth}
    \includegraphics[width=\linewidth - 0.25mm]{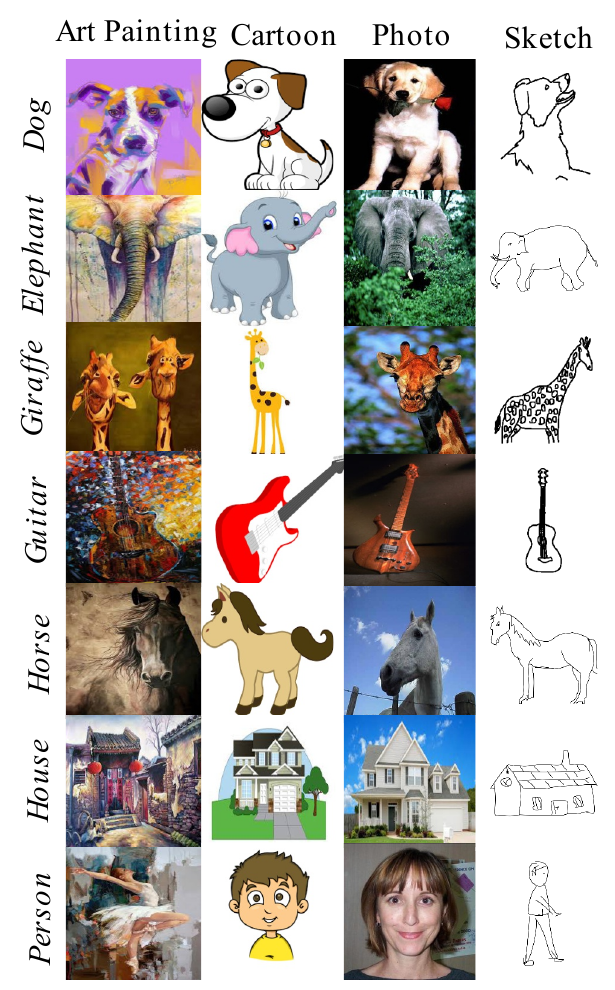}
    \caption{}
    \end{subfigure}
    \begin{subfigure}[t]{.33\linewidth}
    \includegraphics[width=\linewidth - 0.25mm]{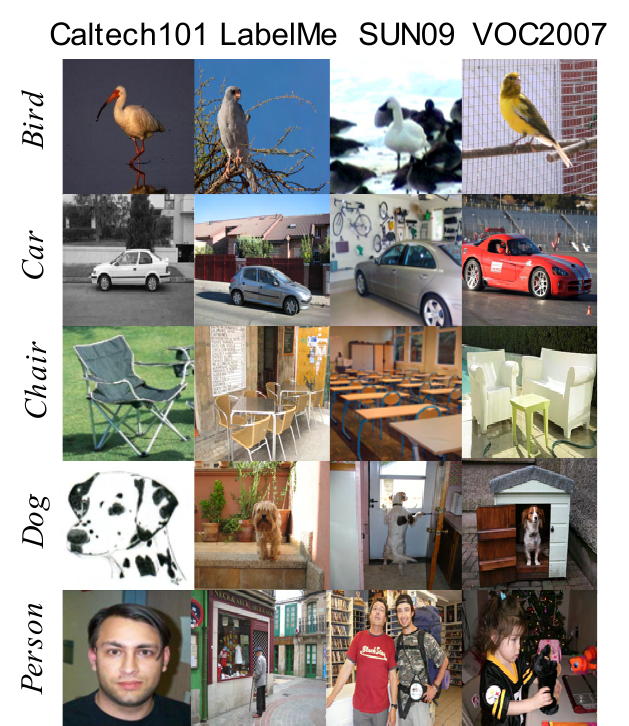}
    \caption{}
    \end{subfigure}
        \begin{subfigure}[t]{.33\linewidth}
    \includegraphics[width=\linewidth - 0.25mm]{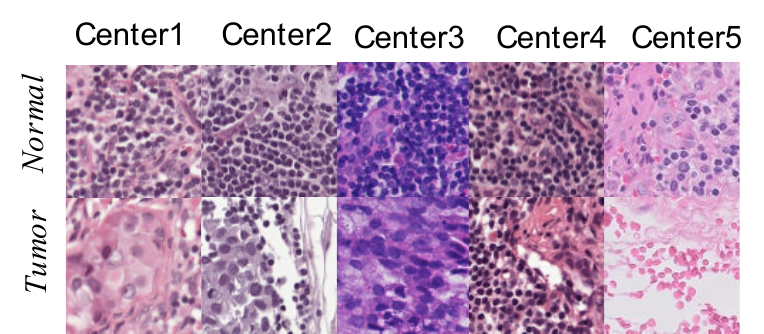}
    \caption{}
    \end{subfigure}
    \caption{\footnotesize \textbf{Samples of DG datasets.} The training data (a) PACS and (b) VLCS (c) Wilds are shown. PACS has four domains \textit{art}~(A), \textit{cartoons}~(C), \textit{photos}~(P), \textit{sketches}~(S). VLCS contains four domains \textit{Caltech101}~(C), \textit{LabelMe}~(L), \textit{SUN09}~(S), \textit{VOC2007}~(V). The WILDS dataset includes data from five different medical centers as domains.} 
    \label{fig:data_vis}
\end{figure*}

We show some images of the datasets in Fig.~\ref{fig:data_vis} to give an intuitive comparison among these image datasets. One can observe that these images have a diverse set of styles, making it very challenging to transfer knowledge from one to another.

\newpage

\section{Additional Experimental Results}
\label{app:exp_results}

\vspace{-2mm}

\begin{figure}[h]
    \begin{subfigure}[h]{.5\linewidth}
    \includegraphics[width=\linewidth - 0.25mm]{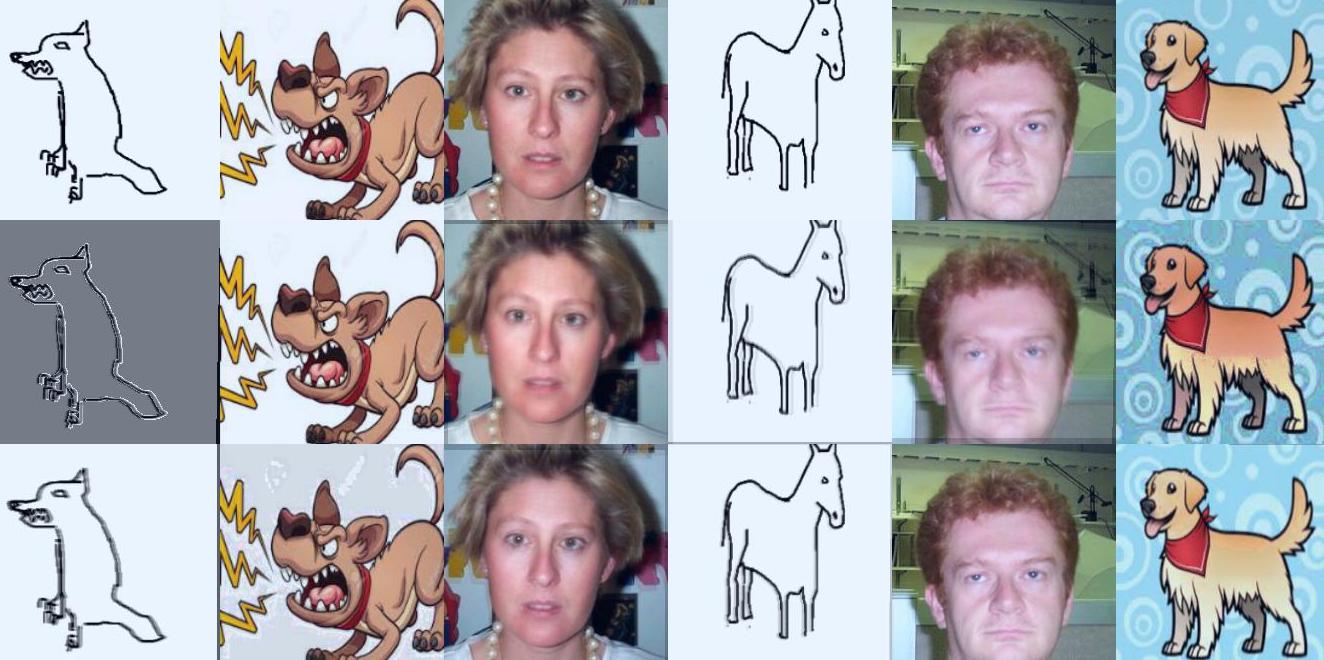} 
    \caption{}
    \end{subfigure}
    \begin{subfigure}[h]{.5\linewidth}
    \includegraphics[width=\linewidth - 0.25mm]{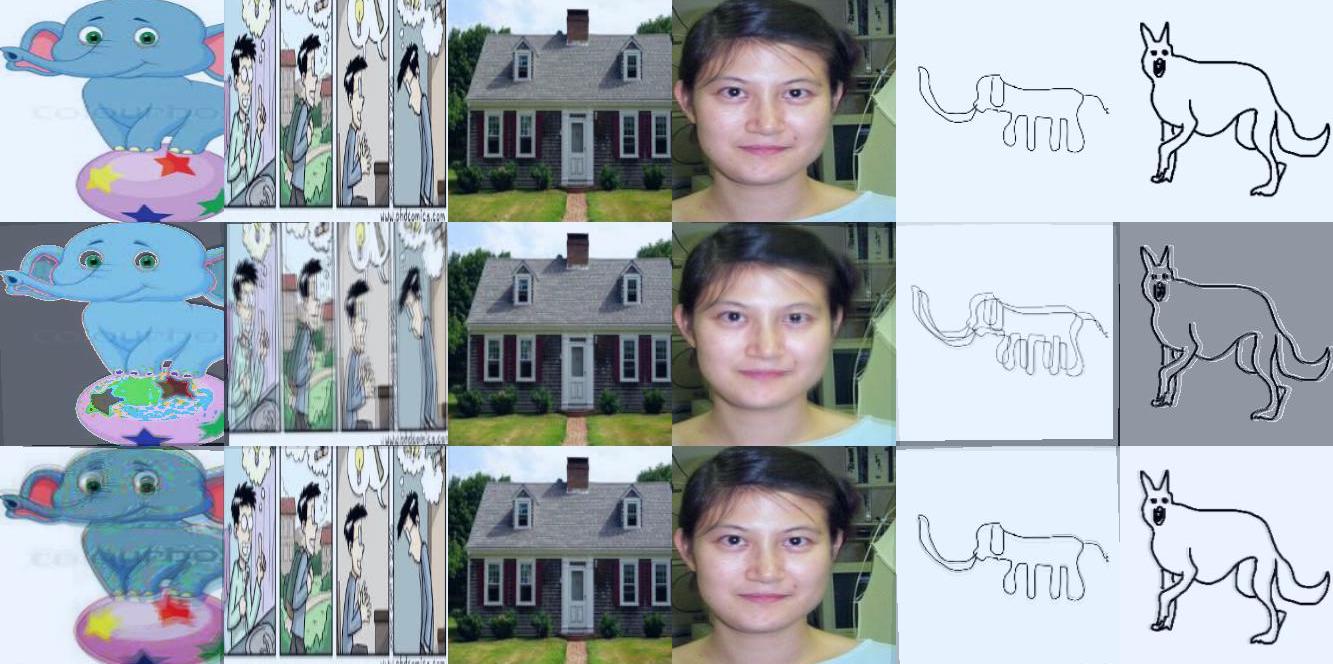}
    \caption{}
    \end{subfigure}
    \caption{\footnotesize \textbf{The augmented samples from AugMix \citep{hendrycks*2020augmix}.} The second and third rows are generated by applying AugMix to the first row.}
    \vspace{-1mm}
    \label{fig:augmix}
\end{figure}

\begin{figure}[h]
    \begin{subfigure}[h]{\linewidth}
    \includegraphics[width=\linewidth - 0.25mm]{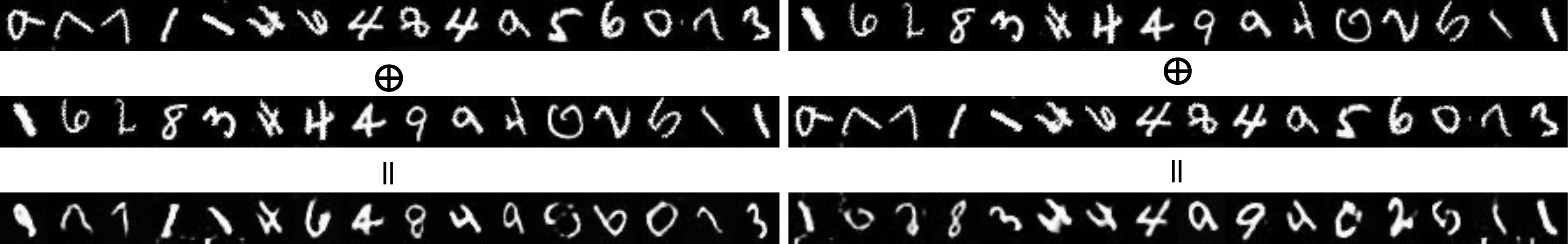} 
    \caption{\footnotesize Rotated MNIST}
    \vspace{4mm}
    \label{fig:add_mnist}
    \end{subfigure}
    \begin{subfigure}[h]{\linewidth}
    \includegraphics[width=\linewidth]{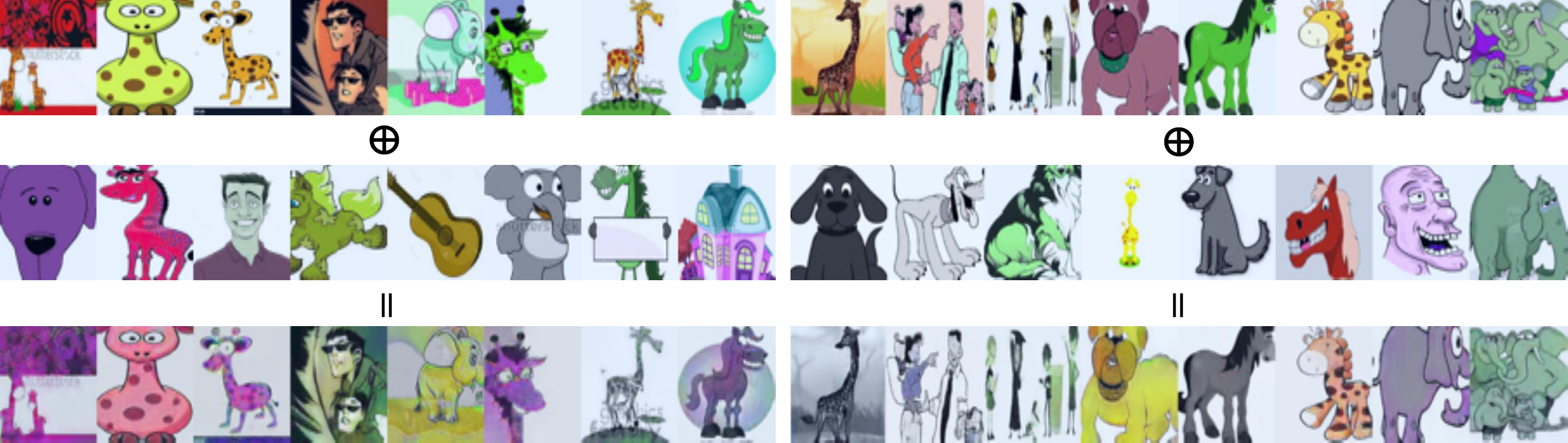}
    \caption{\footnotesize PACS}
    \vspace{4mm}
    \label{fig:add_pacs}
    \end{subfigure}
    \begin{subfigure}[h]{\linewidth}
    \includegraphics[width=\linewidth]{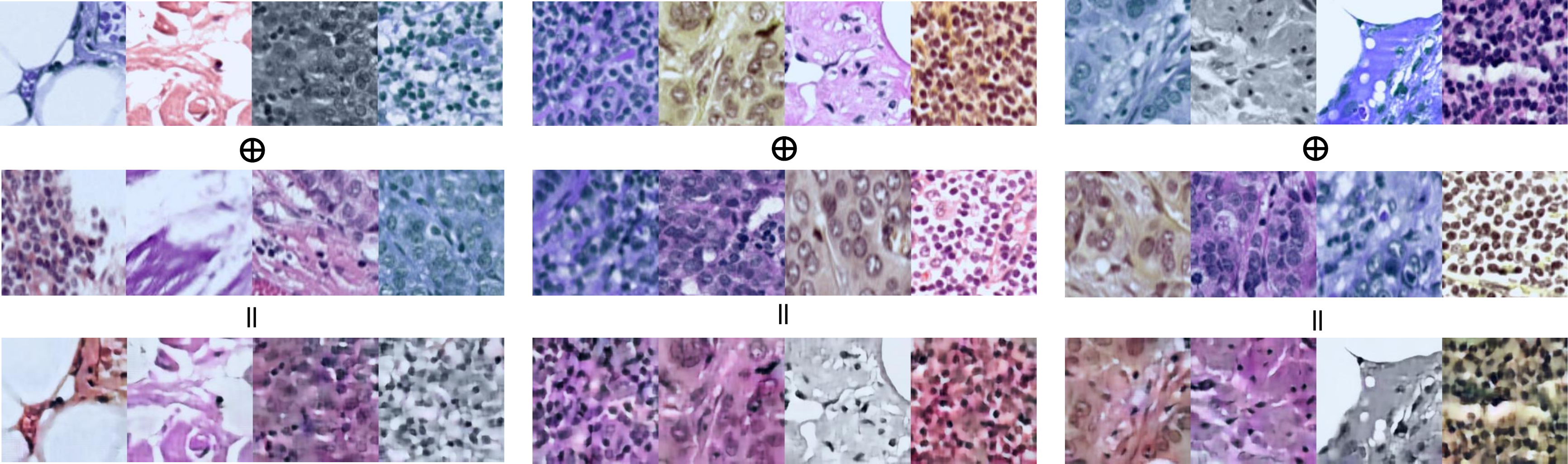}
    \caption{\footnotesize Wilds}
    \vspace{4mm}
    \label{fig:add_wilds}
    \end{subfigure}
    \caption{\footnotesize \textbf{Qualitative disentanglement results on RotatedMNIST, PACS and Wilds.} In every panel, the training data in the first row manifests the semantic factors.} 
    \label{fig:mnist_old}
\end{figure}

\textbf{Qualitative comparison with AugMix.} Looking into the Fig. \ref{fig:augmix}, it is harder for the heuristic-based method AugMix to generate diverse samples via interpolation for training compared with \abbr~as shown in Fig. \ref{fig:main_qual} and Fig. \ref{fig:add_pacs}.

\begin{figure}[h]
    \centering
    \begin{subfigure}[h]{\linewidth}
    \centering
    \includegraphics[width=.94\textwidth]{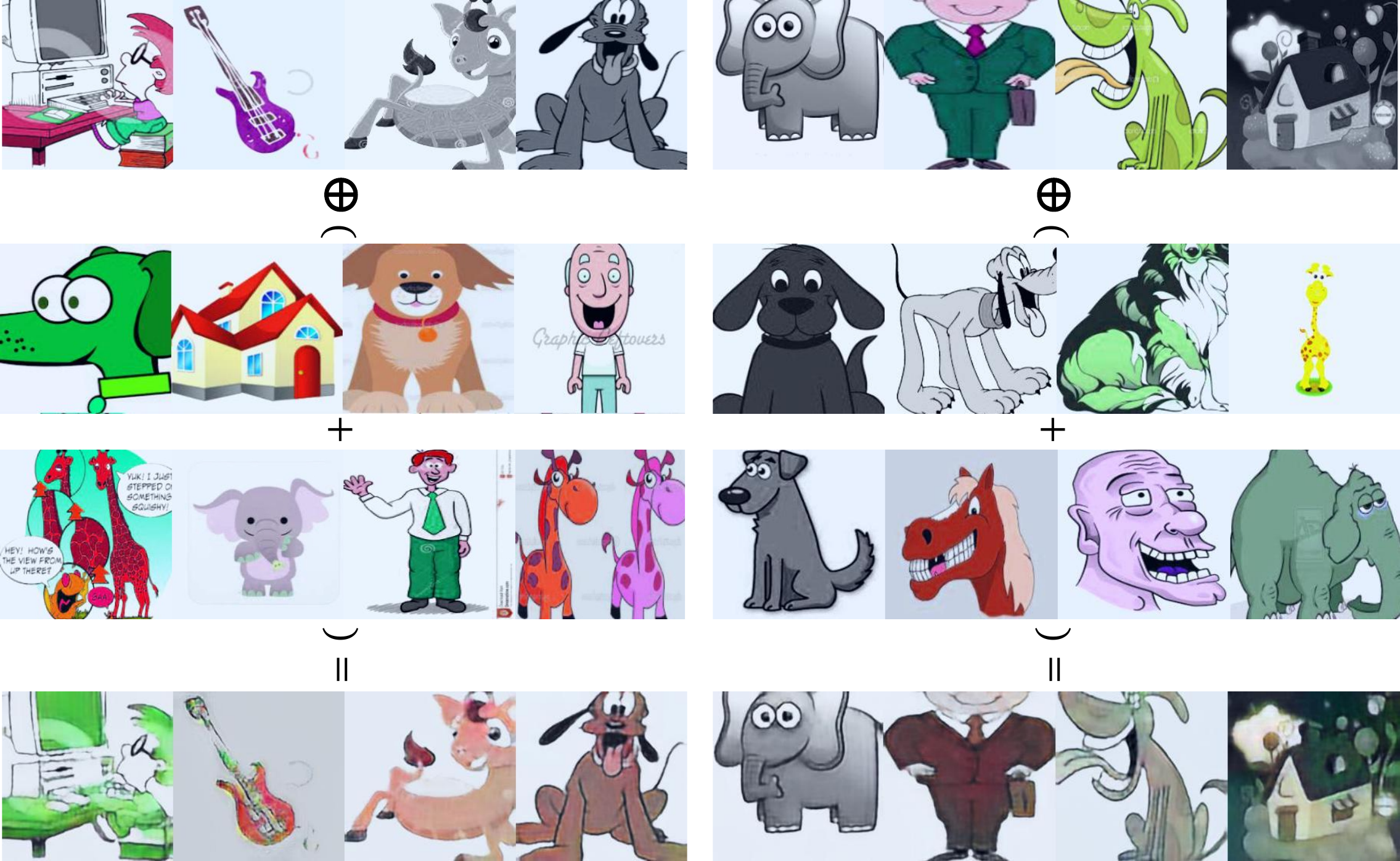}
    \end{subfigure}
    \caption{\footnotesize\textbf{Interpolation via mixing results on PACS.}}
    \label{fig:interpolation_mix}
\end{figure}

\textbf{More qualitative results via interpolation.} Fig. \ref{fig:interpolation_mix} showcases the results of combining the semantic code of one image and the mixture of two variation codes. Results show that the model can generate samples with intermediate variation states.

\textbf{More qualitative results via swapping variation and semantic factors.} We showcase the qualitative results of swapping variation and semantic factors with PACS in Fig. \ref{fig:add_pacs}, MNIST in Fig. \ref{fig:add_mnist}, and WILDS in Fig. \ref{fig:add_wilds}. The results demonstrates the strong disentangled capability of \abbr. 
Some interesting observations are \abbr~learns both intra- (e.g. thickness) and inter-domain (e.g. rotated angle) variations over RotatedMNIST.
\abbr~also maintains semantic information like the color of distinct features across variation-rich data like PACS.

\textbf{Numerical comparison with MBDG.} We quantatively compare \abbr~with MBDG~\citep{robey2021model} as in Table \ref{tab:mbdg}. For a consistent comparison, we run the source code of authors under a test domain validation protocal \citep{gulrajani2021in} on PACS with the results as:

\begin{table*}[h]
\centering
\begin{tabular}{llllll}
 & A & C & P & S & Avg  \\
 \hline
 MBDG & 82.0 $\pm$ 0.0 & 78.4 $\pm$ 0.01 & 93.9 $\pm$ 0.0 & 85.0 $\pm$ 0.0 & 85.8 \\
 \hline
 MBDG$_{Reg}$ & 84.9 $\pm$ 0.0 & 84.9 $\pm$ 0.0 & 93.9 $\pm$ 0.0 & \textbf{85.6 $\pm$ 0.0} & 87.3 \\
 \hline
 DDG & \textbf{88.9 $\pm$ 0.6} & \textbf{85.0 $\pm$ 1.9} & \textbf{97.2 $\pm$ 1.2} & 84.3 $\pm$ 0.7 & \textbf{88.9} \\ 
 \hline
\end{tabular}
\caption{Numerical results for comparing \abbr~and MBDG.}
\label{tab:mbdg}
\end{table*}
Though adopting similar PACCL frameworks and primal-dual algorithms, \abbr~can consistently outperform MBDG and its variant except over domain S. The primary reason behind the clear performance gain of DDG can be that DDG is better at capturing variations within data via random sampling without relying on domain labels. Specifically, parameterizing $h_s, h_v, D$ and constrain them based on disentanglement makes the model more robust to both inter- and intra-domain nuisance factors compared to MBDG that only use a pretrained generator to simulate inter-domain variations. Moreover, the performance gain is also partly due to the three major differences between our approach and MBDG we highlight in the Related Work: First, our upper bound of the parameterization gap is tighter under mild conditions, whereas MBDG requires unrealistic assumptions on the distance metric, i.e., $d(\cdot,\cdot)$ satisfies Lipschitz-like inequality on both arguments, which is stronger than our normal $L_d$ Lipschitzness assumption; second, MBDG consumes additional domain labels, which are prohibitively expensive or even infeasible to obtain in safety-critical applications or those containing sensitive demographics; third, \abbr~enforces invariance constraints via parameterizing semantic and variation encoders, which does not belong to a model-based approach. In contrast, MBDG requires a pre-trained domain transformation model (e.g., CycleGAN) during training, which may result in sub-optimal solutions and parameter inefficiency, while \abbr~is more flexible by treating this as a design choice.

\end{appendix}